\documentclass[twoside,leqno,twocolumn]{article}

\usepackage[letterpaper]{geometry}

\usepackage{ltexpprt}

\usepackage{graphicx}
\usepackage{subcaption}
\graphicspath {{figures/}}
\usepackage{xcolor}
\usepackage{amsmath,amssymb,amsfonts}
\usepackage{cite}
\usepackage[numbers]{natbib}
\usepackage{url}
\usepackage{algorithm}            
\usepackage[noend]{algpseudocode} 
\usepackage{makecell}
\usepackage{tikz}
\usepackage{pgfplots}
\usepackage[title]{appendix}
\newcommand{\N}{\mathcal{N}}
\newcommand{\G}{\mathcal{G}}
\newcommand{\T}{\mathcal{T}}
\newcommand{\W}{\mathcal{W}}

\newcommand{\bbR}{\mathbb{R}}

\newcommand{\bbV}{\mathbb{V}}
\newcommand{\one}{\textbf{1}}
\newcommand{\sdm}{\text{\sc SdM}}
\newcommand{\sdted}{\text{\sc SdTed}}
\newcommand{\mso}{\{\!\!\{}
\newcommand{\msc}{\}\!\!\}}
\DeclareMathOperator*{\argmin}{arg\,min}
\newtheorem{definition}{Definition}

\makeatletter\begin{document}

\title{\Large A Generalized Weisfeiler-Lehman Graph Kernel}

\author{Till Hendrik Schulz \thanks{Dept. of Computer Science, University of Bonn, Germany  Email: \{schulzth, horvath, welke, wrobel\}@cs.uni-bonn.de}
\and Tam\'as Horv\'ath \footnotemark[1] \thanks{Fraunhofer IAIS, Sankt Augustin, Germany} 
\and Pascal Welke \footnotemark[1]
\and Stefan Wrobel \footnotemark[1] \footnotemark[2] 
}

\date{}

\maketitle


\begin{abstract} \small\baselineskip=9pt 
	
	The Weisfeiler-Lehman graph kernels are among the most prevalent graph kernels due to their remarkable time complexity and predictive performance.
	Their key concept is based on an implicit comparison of neighborhood representing trees with respect to equality (i.e., isomorphism).
	This binary valued comparison is, however, arguably too rigid for defining suitable similarity measures over graphs. 
	To overcome this limitation, we propose a generalization of Weisfeiler-Lehman graph kernels which takes into account the similarity between trees rather than equality. 
	We achieve this using a specifically fitted variation of the well-known tree edit distance which can efficiently be calculated.
	We empirically show that our approach significantly outperforms state-of-the-art methods in terms of predictive performance on datasets containing structurally more complex graphs beyond the typically considered molecular graphs.

\end{abstract}

\section{Introduction}

\begin{figure}[t]
	\centering
	\begin{subfigure}[b]{0.15\textwidth}
		\resizebox{0.95\textwidth}{!}{\begingroup
  \makeatletter
  \providecommand\color[2][]{
    \errmessage{(Inkscape) Color is used for the text in Inkscape, but the package 'color.sty' is not loaded}
    \renewcommand\color[2][]{}
  }
  \providecommand\transparent[1]{
    \errmessage{(Inkscape) Transparency is used (non-zero) for the text in Inkscape, but the package 'transparent.sty' is not loaded}
    \renewcommand\transparent[1]{}
  }
  \providecommand\rotatebox[2]{#2}
  \newcommand*\fsize{\dimexpr\f@size pt\relax}
  \newcommand*\lineheight[1]{\fontsize{\fsize}{#1\fsize}\selectfont}
  \ifx\svgwidth\undefined
    \setlength{\unitlength}{126.65969008bp}
    \ifx\svgscale\undefined
      \relax
    \else
      \setlength{\unitlength}{\unitlength * \real{\svgscale}}
    \fi
  \else
    \setlength{\unitlength}{\svgwidth}
  \fi
  \global\let\svgwidth\undefined
  \global\let\svgscale\undefined
  \makeatother
  \begin{picture}(1,0.93205924)
    \lineheight{1}
    \setlength\tabcolsep{0pt}
    \put(0,0){\includegraphics[width=\unitlength,page=1]{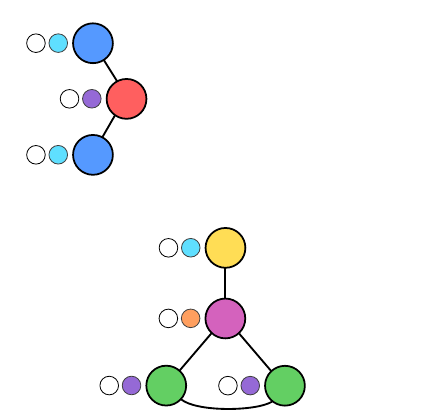}}
    \put(0.55032421,0.90239273){\color[rgb]{0,0,0}\makebox(0,0)[lt]{\lineheight{1.25}\smash{\begin{tabular}[t]{l}$G_2:$\end{tabular}}}}
    \put(-0.00279761,0.90645292){\color[rgb]{0,0,0}\makebox(0,0)[lt]{\lineheight{1.25}\smash{\begin{tabular}[t]{l}$G_1:$\end{tabular}}}}
    \put(0.28173189,0.42212583){\color[rgb]{0,0,0}\makebox(0,0)[lt]{\lineheight{1.25}\smash{\begin{tabular}[t]{l}$G_3:$\end{tabular}}}}
    \put(0,0){\includegraphics[width=\unitlength,page=2]{WL_colored_graphs.pdf}}
  \end{picture}
\endgroup
 }
		\caption{}
		\label{fig:workflow_graphs}
	\end{subfigure}
	\hfill
	\begin{subfigure}[b]{0.15\textwidth}
		\resizebox{0.95\textwidth}{!}{\begingroup
  \makeatletter
  \providecommand\color[2][]{
    \errmessage{(Inkscape) Color is used for the text in Inkscape, but the package 'color.sty' is not loaded}
    \renewcommand\color[2][]{}
  }
  \providecommand\transparent[1]{
    \errmessage{(Inkscape) Transparency is used (non-zero) for the text in Inkscape, but the package 'transparent.sty' is not loaded}
    \renewcommand\transparent[1]{}
  }
  \providecommand\rotatebox[2]{#2}
  \newcommand*\fsize{\dimexpr\f@size pt\relax}
  \newcommand*\lineheight[1]{\fontsize{\fsize}{#1\fsize}\selectfont}
  \ifx\svgwidth\undefined
    \setlength{\unitlength}{126.17960051bp}
    \ifx\svgscale\undefined
      \relax
    \else
      \setlength{\unitlength}{\unitlength * \real{\svgscale}}
    \fi
  \else
    \setlength{\unitlength}{\svgwidth}
  \fi
  \global\let\svgwidth\undefined
  \global\let\svgscale\undefined
  \makeatother
  \begin{picture}(1,0.78091955)
    \lineheight{1}
    \setlength\tabcolsep{0pt}
    \put(0.65926057,0.75749078){\color[rgb]{0,0,0}\makebox(0,0)[lt]{\lineheight{1.25}\smash{\begin{tabular}[t]{l}$T_2:$\end{tabular}}}}
    \put(0,0){\includegraphics[width=\unitlength,page=1]{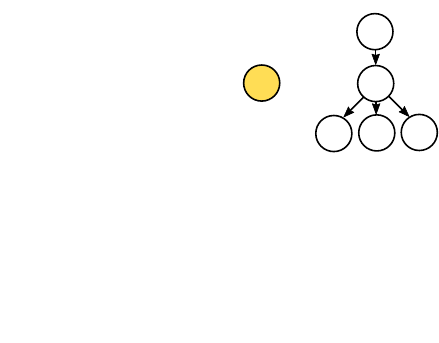}}
    \put(0.70075133,0.58690535){\color[rgb]{0,0,0}\makebox(0,0)[lt]{\lineheight{1.25}\smash{\begin{tabular}[t]{l}$\equiv$\end{tabular}}}}
    \put(0.113014,0.75763475){\color[rgb]{0,0,0}\makebox(0,0)[lt]{\lineheight{1.25}\smash{\begin{tabular}[t]{l}$T_1:$\end{tabular}}}}
    \put(0,0){\includegraphics[width=\unitlength,page=2]{unfolded_trees.pdf}}
    \put(0.14444453,0.57765804){\color[rgb]{0,0,0}\makebox(0,0)[lt]{\lineheight{1.25}\smash{\begin{tabular}[t]{l}$\equiv$\end{tabular}}}}
    \put(0,0){\includegraphics[width=\unitlength,page=3]{unfolded_trees.pdf}}
    \put(0.2897659,0.15544397){\color[rgb]{0,0,0}\makebox(0,0)[lt]{\lineheight{1.25}\smash{\begin{tabular}[t]{l}$\equiv$\end{tabular}}}}
    \put(0.39091639,0.30506343){\color[rgb]{0,0,0}\makebox(0,0)[lt]{\lineheight{1.25}\smash{\begin{tabular}[t]{l}$T_3:$\end{tabular}}}}
    \put(0,0){\includegraphics[width=\unitlength,page=4]{unfolded_trees.pdf}}
  \end{picture}
\endgroup
 }
		\caption{}
		\label{fig:workflow_unfolding_trees}
	\end{subfigure}
	\hfill
	\begin{subfigure}[b]{0.15\textwidth}
		\resizebox{0.8\textwidth}{!}{\begingroup
  \makeatletter
  \providecommand\color[2][]{
    \errmessage{(Inkscape) Color is used for the text in Inkscape, but the package 'color.sty' is not loaded}
    \renewcommand\color[2][]{}
  }
  \providecommand\transparent[1]{
    \errmessage{(Inkscape) Transparency is used (non-zero) for the text in Inkscape, but the package 'transparent.sty' is not loaded}
    \renewcommand\transparent[1]{}
  }
  \providecommand\rotatebox[2]{#2}
  \newcommand*\fsize{\dimexpr\f@size pt\relax}
  \newcommand*\lineheight[1]{\fontsize{\fsize}{#1\fsize}\selectfont}
  \ifx\svgwidth\undefined
    \setlength{\unitlength}{133.12111081bp}
    \ifx\svgscale\undefined
      \relax
    \else
      \setlength{\unitlength}{\unitlength * \real{\svgscale}}
    \fi
  \else
    \setlength{\unitlength}{\svgwidth}
  \fi
  \global\let\svgwidth\undefined
  \global\let\svgscale\undefined
  \makeatother
  \begin{picture}(1,1.08159525)
    \lineheight{1}
    \setlength\tabcolsep{0pt}
    \put(0,0){\includegraphics[width=\unitlength,page=1]{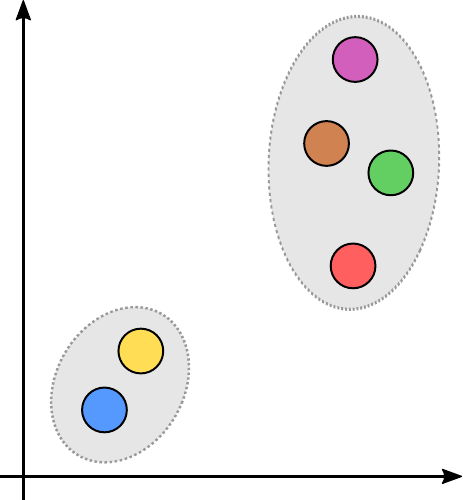}}
  \end{picture}
\endgroup
 }
		\caption{}
		\label{fig:workflow_clustering}
	\end{subfigure}
	\caption{ 
		(a) depicts (initially unlabeled) graphs where vertices are labeled with the first two Weisfeiler-Lehman labels (colored). 
		(b) shows the rooted unfolding trees corresponding to the blue, yellow and pink WL-labels, each representing a neighborhood. 
		$T_1$ (blue) differs from $T_2$ (yellow) by only a single vertex while it differs from $T_3$ (pink) by significantly more. 
		The tree edit distance between $T_1$ and $T_2$ is therefore much smaller than that between $T_1$ and $T_3$. 
		(c) conceptually visualizes the latent space representing the pairwise tree edit distances between unfolding trees. 
		Clusterings in this space identify groups of pairwise similar unfolding trees. }
	\label{fig:workflow}
\end{figure}

Since Haussler's pioneer work~\citep{Haussler99} 
on convolution kernels over discrete structures, {\em graph kernels} have become one of the most common tools for learning with graphs. 
They gained major popularity by enabling the application of kernel methods.
For example, excellent predictive performance can be obtained by combining graph kernels with support vector machines.
One prominent family of graph kernels is the {\em Weisfeiler-Lehman kernel} framework~\citep{shervashidze2011weisfeiler}.
Kernels in this family are based on the idea of the {\em Weisfeiler-Lehman isomorphism test}~\citep{weisfeilerlehman}, which iteratively relabels vertices by propagating neighborhood information.
Each such label implicitly corresponds to a rooted tree, called {\em unfolding tree} (see Fig. \ref{fig:workflow_unfolding_trees}). 
For space limitations, we limit the scope of this work to the most established member, the \emph{Weisfeiler-Lehman subtree kernel}.
However, we note that the generality of our approach allows its application to \emph{all} Weisfeiler-Lehman graph kernels.

Despite their distinguished speed, Weisfeiler-Lehman graph kernels are conceptually limited to comparing labels, or equivalently, unfolding trees w.r.t. \emph{equality}. 
While this comparison is extremely well-suited for graph isomorphism tests, it is arguably too restrictive for defining \emph{similarities}, in particular, graph kernels.
As an example, consider the unfolding trees depicted in Fig. \ref{fig:workflow_unfolding_trees}. 
While $T_1$ visibly resembles $T_2$ much more than it resembles $T_3$, the Weisfeiler-Lehman kernel~\citep{shervashidze2011weisfeiler} simply treats them all as unequal and is thus unable to \textit{quantify} the apparent difference among the pairwise similarities between the unfolding trees.

Motivated by these considerations, 
we {\em relax} the above strictness by proposing a method which compares Weisfeiler-Lehman labels, or equivalently unfolding trees, with regard to a much {\em finer} similarity measure than the binary valued one. 
More precisely, we employ a similarity between Weisfeiler-Lehman labels based on the concept of {\em tree edit distances} between their respective unfolding trees.
These kind of distances provide a natural comparison for trees. 
On an abstract level, they are defined by the minimum cumulative cost of {\em edit operations} needed to transform
one tree into another.
Since in this work we deal with unfolding trees, we define a variant of the tree edit distance specific to this special type of rooted trees. 
We show that in contrast to more general tree edit distances, this distance can in fact be efficiently calculated.

The key concept of our {\em relaxed Weisfeiler-Lehman subtree kernel} is to identify groups of {\em similar} Weisfeiler-Lehman labels by {\em clustering} (visualized in Fig. \ref{fig:workflow_clustering}). 
The elements within a cluster are then treated as {\em identical} labels. 
That is, we generalize the ordinary Weisfeiler-Lehman kernel~\citep{shervashidze2011weisfeiler} by regarding two unfolding trees equivalent if they belong to the same cluster, i.e., if they have a {\em small} distance to each other.
In this way, the ordinary Weisfeiler-Lehman kernel is the {\em special} case where labels are considered equivalent only if they have distance zero. 
For partitioning the Weisfeiler-Lehman labels, we use {\em Wasserstein $k$-means} clustering~\citep{DBLP:journals/eswa/IrpinoVC14}. 
This choice is motivated by our result that the tree edit distance between unfolding trees can in fact be reformulated in terms of the Wasserstein distance. 

We have empirically evaluated the predictive performance of our relaxed
Weisfeiler-Lehman kernel on a set of real-world datasets. 
Our experimental results clearly show that our approach considerably outperforms state-of-the-art kernels (including the ordinary Weisfeiler-Lehman subtree kernel) on datasets containing dense and structurally diverse graphs.

\paragraph{Related Work}
\label{sec:rel_work}

While conventional graph kernels define similarity in terms of mutual substructures such as walks \cite{gaertner2003}, paths \cite{borgwardt2005}, small subgraphs \cite{DBLP:journals/jmlr/Shervashidze2009graphlet} or subtrees \cite{shervashidze2011weisfeiler}, recent work has moved away from solely counting equivalent substructures. 
For example, \cite{kriege16} introduces a kernel which computes an optimal assignment between vertices. 
Similarly, in \cite{NIPS2019_WassersteinWeisfeilerLehman} the authors employ the concept of optimal transportation as a form of ``soft-matching'' on vertices. 
Both methods measure similarity between vertices based on variants of Weisfeiler-Lehman induced similarity.
However, their vertex matching abilities are ultimately still limited by the rigid Weisfeiler-Lehman label refinement method. 
 
The concept of comparing vertices by a finer similarity measure than the equivalence of their neighborhoods can also be found in \cite{martino2012}. 
In that work, Martino et al. represent neighborhoods by rooted directed acyclic trees (DAG) and define a kernel over these DAGs which reflects their similarity. 
The difference to our approach is twofold. 
Firstly, compared to Weisfeiler-Lehman labels, the DAGs describe structurally different representations of neighborhoods .
Secondly, while the authors in \cite{martino2012} compute similarities by applying tree kernels (e.g. \cite{smola2002}) on sets of trees extracted from the DAGs, we employ the concept of tree edit distances as similarity measure.

The rest of the paper is organized as follows. 
We 
collect the necessary notions in Sect.~\ref{sec:preliminaries}, discuss the tree edit distance in Sect.~\ref{sec:editdistance}, and present our graph kernel in Sect.~\ref{sec:kernel}. 
We report the empirical results in Sect.~\ref{sec:experiments} and conclude in Sect.~\ref{sec:conclusion}.

\section{Preliminaries}\label{sec:preliminaries}

	\paragraph*{Graphs}
	An \emph{(undirected) graph} $G = (V,E,\ell)$ consists of a finite set $V$ of vertices, a set $E \subseteq \{ X \subseteq V : |X| = 2 \}$ of edges, and a label function $\ell: V \to \Sigma$ for some finite alphabet $\Sigma$.
	When $G$ is clear from the context, we use $n := |V|$ and $m := |E|$.
	For $v \in V$, $\N(v)$ is the set of \emph{neighbors} of node $v$.
	Two graphs $G,G'$ are \emph{isomorphic}, denoted $G \equiv G'$, if there exists a bijective function between the vertices of $G$ and those of $G'$ preserving all edges and labels in both directions.
	A \emph{(rooted) tree} is a connected graph $T=(V,E)$ that has $n-1$ edges and a \emph{root} $r(T) \in V$. 
	For any $v \in V\setminus \{r(T)\}$, $par(v)$ is the \emph{parent} of $v$, i.e., the unique neighbor of $v$ on the path to $r(T)$; accordingly, the children of $v$ are all vertices that have $v$ as parent.
	The \emph{subtree rooted in $v$}, denoted $T[v]$, is the subgraph of $T$ that is rooted at $v$ and induced by all descendants of $v$. 
	$F(v)$ then denotes the set of subtrees rooted at the children of $v$. 
	
	\paragraph*{Tree edit distance}
	Let $\bot \not \in \Sigma$ be a special \emph{blank} symbol.
	For $\Sigma^{\bot} = \Sigma \cup \{\bot\}$ we define a \emph{cost function} $\gamma: \Sigma^{\bot} \times \Sigma^{\bot} \rightarrow \bbR$ and require $\gamma$ to be a metric.
	An \emph{edit script} or \emph{edit sequence} from a tree $T$ into a tree $T'$ is a sequence of edit operations turning $T$ into $T'$.
	An edit operation can (i) \emph{relabel} a single node $v$, (ii) \emph{delete} $v$ and connect all its children to the parent of $v$, or (iii) \emph{insert} a single node $w$ between $v$ and a subset of $v$'s children.
	The cost of such edits is defined by $\gamma$; relabeling $v$ from $a$ to $b$ costs $\gamma(a,b)$ and adding or deleting $v$ costs $\gamma(\ell(v), \bot)$.
	An edit script between $T$ and $T'$ of minimum cost is called \emph{optimal} and its cost is called \emph{tree edit distance}.
	It is a metric if $\gamma$ is a metric.
	
	\paragraph*{Wasserstein distance}
	Given two vectors $x \in \bbR^n$ and $x' \in \mathbb{R}^{n'}$ with $|x|_1=|x'|_1$ and a cost matrix $C^{n \times n'}$ containing pairwise distances between entries of $x$ and $x'$, the \emph{Wasserstein distance} 
	is defined by 
	\[ \W^C(x,x') = \min_{T \in \mathcal{T}(x,x')} \langle T,C \rangle \]
	with $\mathcal{T}(x,x') \subseteq \mathbb{R}^{n \times n'}$ and $T\one_{n'}=x$, $\one_n^T T=x'$ for all $T \in \mathcal{T}(x,x')$, where $\langle .,. \rangle$ is the Frobenius inner product.
	A $T \in \mathcal{T}(x,x')$ is called \emph{transport matrix} and a minimizer of the above is called \emph{optimal} transport matrix.
	If the cost matrix is defined by a metric, then the Wasserstein distance is a metric.
	For a set of vectors $x_1,...,x_n \in \bbR^n$ and a cost matrix $C^{n \times n}$, we define the \emph{barycenter} as $\argmin_c \sum_{i \in [n]} \W^C(x_i,c)$.

\section{The Weisfeiler-Lehman Tree Edit Distance}
\label{sec:editdistance}

	In this section, we briefly recap the Weisfeiler-Lehman vertex relabeling method \citep{weisfeilerlehman} and define a distance function on Weisfeiler-Lehman labels.
	We give an algorithm computing this distance and prove that it can be efficiently calculated.
	
	\subsection{The Weisfeiler-Lehman method}
	The \emph{Weisfeiler-Lehman} (WL) method \citep{weisfeilerlehman} was originally designed to decide isomorphism between graphs with one-sided error.
	Its key idea is to iteratively refine a partitioning of the vertex set by compressing the labels of each node and its neighbors into a new label.
	This is done by concatenating a node's label and its ordered (multi-)set of neighbor labels and subsequently hashing it to a new label by a perfect hash function.
	Thus, with each iteration, labels incorporate increasingly large substructures.  
	The injectivity of the hash function ensures that different sorted lists of labels cannot be mapped to the same (new) label.
	
	More precisely, let $G=(V,E,\ell_0)$ be a graph with initial vertex label function $\ell_0:V \rightarrow \Sigma_0$, where $\Sigma_0$ is the alphabet of the original vertex labels. 
	In case of unlabeled graphs, we assume all vertices to have the same mutual label.
	Assuming that there is a total order on alphabet $\Sigma_i$ for all $i \geq 0$, the Weisfeiler-Lehman algorithm recursively computes the new label of $v$ in iteration $i+1$ by 
	\[ \ell_{i+1}(v) = f_\#(\ell_i(v), [ \ell_i(u): u \in \N(v) ]) \in \Sigma_{i+1} \]
	for all vertices $v$, where the list of labels in the second argument of $f_\#$ is sorted by the total order on $\Sigma_i$ and $f_\#:\Sigma_i \times \Sigma_i^* \rightarrow \Sigma_{i+1}$ is a perfect (i.e., injective) hash function.
	Two graphs $G,G'$ are not isomorphic if the corresponding multisets $\mso \ell_{i}(v): v \in V(G) \msc$ and $\mso \ell_{i}(v'): v' \in V(G') \msc$ are different for some $i \in \mathbb{N}$; otherwise they may or may not be isomorphic. 
	
	Shervashidze et al. \cite{shervashidze2011weisfeiler} employed the Weisfeiler-Lehman method to define a family of parameterized kernels measuring the similarity between graphs based on their relabeled versions. 
	For a graph $G=(V,E,\ell_0)$ they consider the sequence of WL-graphs $G_0,G_1,...,G_h$ with $G_i=(V,E,\ell_i)$, where $h$ is the number of performed WL iterations.
	The Weisfeiler-Lehman kernel of depth $h$ for two graphs $G,G'$, given some base graph kernel $k$, is then defined as 
	\[ k^h_{WL}(G,G') = \sum_{i=0,...h}k(G_i,G'_i) \enspace . \]
	In other words, the kernel $k$ is applied to $G,G'$ for all labeling functions $\ell_i$  ($0 \leq i \leq h$) and the $h+1$ values obtained are subsequently summed up.
	We note that each component $k(G_i,G'_i)$ in $k^h_{WL}(G,G')$ can be assigned a non-negative real weight $\alpha_i$.
	This allows e.g. to emphasize larger substructures (i.e., labels in higher iterations contribute more to the overall similarity).
	While the base kernel $k$ can be an arbitrary positive semi-definite kernel on graphs, for space limitations we focus on the {\em subtree kernel}~\cite{shervashidze2011weisfeiler} which employs the base kernel 
	\[ k(G_i,G_i')=\sum_{v \in V}\sum_{v' \in \ V'} \delta(\ell_i(v), \ell_i(v'))\enspace ,\] 
	where $\delta$ is the Kronecker delta.
	Thus, $k^h_{WL}$ simply counts the pairs of matching labels of all WL-iterations.
	With complexity $O(hm)$, where $m$ is the number of edges, the WL subtree kernel is highly efficient and has proven to provide state-of-the-art results on a broad range of datasets. 
	
\begin{figure}[t]
		\centering
		\begin{subfigure}[b]{0.4\textwidth}
			\resizebox{0.9\textwidth}{!}{\begingroup
  \makeatletter
  \providecommand\color[2][]{
    \errmessage{(Inkscape) Color is used for the text in Inkscape, but the package 'color.sty' is not loaded}
    \renewcommand\color[2][]{}
  }
  \providecommand\transparent[1]{
    \errmessage{(Inkscape) Transparency is used (non-zero) for the text in Inkscape, but the package 'transparent.sty' is not loaded}
    \renewcommand\transparent[1]{}
  }
  \providecommand\rotatebox[2]{#2}
  \newcommand*\fsize{\dimexpr\f@size pt\relax}
  \newcommand*\lineheight[1]{\fontsize{\fsize}{#1\fsize}\selectfont}
  \ifx\svgwidth\undefined
    \setlength{\unitlength}{272.63429657bp}
    \ifx\svgscale\undefined
      \relax
    \else
      \setlength{\unitlength}{\unitlength * \real{\svgscale}}
    \fi
  \else
    \setlength{\unitlength}{\svgwidth}
  \fi
  \global\let\svgwidth\undefined
  \global\let\svgscale\undefined
  \makeatother
  \begin{picture}(1,0.2640913)
    \lineheight{1}
    \setlength\tabcolsep{0pt}
    \put(0.00190036,0.2474225){\color[rgb]{0,0,0}\makebox(0,0)[lt]{\lineheight{1.25}\smash{\begin{tabular}[t]{l}$G:$\end{tabular}}}}
    \put(0,0){\includegraphics[width=\unitlength,page=1]{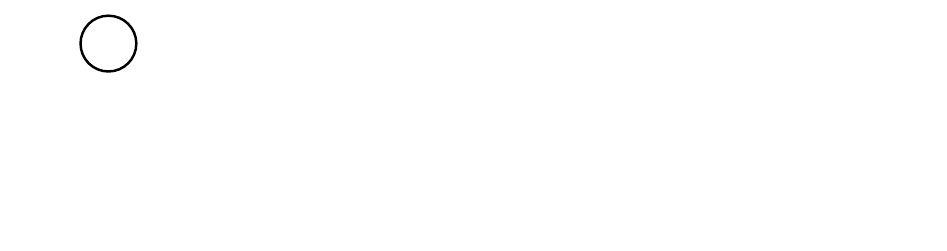}}
    \put(0.1066434,0.20730835){\color[rgb]{0,0,0}\makebox(0,0)[lt]{\lineheight{1.25}\smash{\begin{tabular}[t]{l}$1$\end{tabular}}}}
    \put(0,0){\includegraphics[width=\unitlength,page=2]{unfolding_example_a.pdf}}
    \put(0.02288579,0.15521472){\color[rgb]{0,0,0}\makebox(0,0)[lt]{\lineheight{1.25}\smash{\begin{tabular}[t]{l}$2$\end{tabular}}}}
    \put(0,0){\includegraphics[width=\unitlength,page=3]{unfolding_example_a.pdf}}
    \put(0.19197021,0.15509182){\color[rgb]{0,0,0}\makebox(0,0)[lt]{\lineheight{1.25}\smash{\begin{tabular}[t]{l}$3$\end{tabular}}}}
    \put(0,0){\includegraphics[width=\unitlength,page=4]{unfolding_example_a.pdf}}
    \put(0.0229097,0.07258905){\color[rgb]{0,0,0}\makebox(0,0)[lt]{\lineheight{1.25}\smash{\begin{tabular}[t]{l}$4$\end{tabular}}}}
    \put(0,0){\includegraphics[width=\unitlength,page=5]{unfolding_example_a.pdf}}
    \put(0.19199408,0.07246616){\color[rgb]{0,0,0}\makebox(0,0)[lt]{\lineheight{1.25}\smash{\begin{tabular}[t]{l}$5$\end{tabular}}}}
    \put(0,0){\includegraphics[width=\unitlength,page=6]{unfolding_example_a.pdf}}
    \put(0.10670121,0.02024849){\color[rgb]{0,0,0}\makebox(0,0)[lt]{\lineheight{1.25}\smash{\begin{tabular}[t]{l}$6$\end{tabular}}}}
    \put(0,0){\includegraphics[width=\unitlength,page=7]{unfolding_example_a.pdf}}
    \put(0.37780413,0.11380422){\color[rgb]{0,0,0}\makebox(0,0)[lt]{\lineheight{1.25}\smash{\begin{tabular}[t]{l}$2$\end{tabular}}}}
    \put(0,0){\includegraphics[width=\unitlength,page=8]{unfolding_example_a.pdf}}
    \put(0.34629185,0.02015737){\color[rgb]{0,0,0}\makebox(0,0)[lt]{\lineheight{1.25}\smash{\begin{tabular}[t]{l}$1$\end{tabular}}}}
    \put(0,0){\includegraphics[width=\unitlength,page=9]{unfolding_example_a.pdf}}
    \put(0.41219146,0.02003469){\color[rgb]{0,0,0}\makebox(0,0)[lt]{\lineheight{1.25}\smash{\begin{tabular}[t]{l}$4$\end{tabular}}}}
    \put(0,0){\includegraphics[width=\unitlength,page=10]{unfolding_example_a.pdf}}
    \put(0.51625516,0.11380422){\color[rgb]{0,0,0}\makebox(0,0)[lt]{\lineheight{1.25}\smash{\begin{tabular}[t]{l}$3$\end{tabular}}}}
    \put(0,0){\includegraphics[width=\unitlength,page=11]{unfolding_example_a.pdf}}
    \put(0.48474286,0.02015737){\color[rgb]{0,0,0}\makebox(0,0)[lt]{\lineheight{1.25}\smash{\begin{tabular}[t]{l}$1$\end{tabular}}}}
    \put(0,0){\includegraphics[width=\unitlength,page=12]{unfolding_example_a.pdf}}
    \put(0.55064249,0.02003469){\color[rgb]{0,0,0}\makebox(0,0)[lt]{\lineheight{1.25}\smash{\begin{tabular}[t]{l}$5$\end{tabular}}}}
    \put(0,0){\includegraphics[width=\unitlength,page=13]{unfolding_example_a.pdf}}
    \put(0.68827383,0.11371054){\color[rgb]{0,0,0}\makebox(0,0)[lt]{\lineheight{1.25}\smash{\begin{tabular}[t]{l}$4$\end{tabular}}}}
    \put(0,0){\includegraphics[width=\unitlength,page=14]{unfolding_example_a.pdf}}
    \put(0.62272272,0.02019844){\color[rgb]{0,0,0}\makebox(0,0)[lt]{\lineheight{1.25}\smash{\begin{tabular}[t]{l}$1$\end{tabular}}}}
    \put(0,0){\includegraphics[width=\unitlength,page=15]{unfolding_example_a.pdf}}
    \put(0.68862234,0.02007575){\color[rgb]{0,0,0}\makebox(0,0)[lt]{\lineheight{1.25}\smash{\begin{tabular}[t]{l}$2$\end{tabular}}}}
    \put(0,0){\includegraphics[width=\unitlength,page=16]{unfolding_example_a.pdf}}
    \put(0.75458633,0.02003469){\color[rgb]{0,0,0}\makebox(0,0)[lt]{\lineheight{1.25}\smash{\begin{tabular}[t]{l}$6$\end{tabular}}}}
    \put(0,0){\includegraphics[width=\unitlength,page=17]{unfolding_example_a.pdf}}
    \put(0.89361171,0.11371043){\color[rgb]{0,0,0}\makebox(0,0)[lt]{\lineheight{1.25}\smash{\begin{tabular}[t]{l}$5$\end{tabular}}}}
    \put(0,0){\includegraphics[width=\unitlength,page=18]{unfolding_example_a.pdf}}
    \put(0.8280606,0.02019844){\color[rgb]{0,0,0}\makebox(0,0)[lt]{\lineheight{1.25}\smash{\begin{tabular}[t]{l}$1$\end{tabular}}}}
    \put(0,0){\includegraphics[width=\unitlength,page=19]{unfolding_example_a.pdf}}
    \put(0.89396026,0.02007575){\color[rgb]{0,0,0}\makebox(0,0)[lt]{\lineheight{1.25}\smash{\begin{tabular}[t]{l}$3$\end{tabular}}}}
    \put(0,0){\includegraphics[width=\unitlength,page=20]{unfolding_example_a.pdf}}
    \put(0.95992421,0.02003468){\color[rgb]{0,0,0}\makebox(0,0)[lt]{\lineheight{1.25}\smash{\begin{tabular}[t]{l}$6$\end{tabular}}}}
    \put(0,0){\includegraphics[width=\unitlength,page=21]{unfolding_example_a.pdf}}
    \put(0.62625653,0.20718514){\color[rgb]{0,0,0}\makebox(0,0)[lt]{\lineheight{1.25}\smash{\begin{tabular}[t]{l}$1$\end{tabular}}}}
    \put(0,0){\includegraphics[width=\unitlength,page=22]{unfolding_example_a.pdf}}
    \put(0.13567236,0.24600375){\color[rgb]{0,0,0}\makebox(0,0)[lt]{\lineheight{1.25}\smash{\begin{tabular}[t]{l}$v$\end{tabular}}}}
    \put(0.33625675,0.24686522){\color[rgb]{0,0,0}\makebox(0,0)[lt]{\lineheight{1.25}\smash{\begin{tabular}[t]{l}$T^2(G,v):$\end{tabular}}}}
  \end{picture}
\endgroup
 }
			\caption{}
			\label{fig:unfolding_example_a}
		\end{subfigure}
		\hfill
		\begin{subfigure}[b]{0.4\textwidth}
			\resizebox{0.9\textwidth}{!}{\begingroup
  \makeatletter
  \providecommand\color[2][]{
    \errmessage{(Inkscape) Color is used for the text in Inkscape, but the package 'color.sty' is not loaded}
    \renewcommand\color[2][]{}
  }
  \providecommand\transparent[1]{
    \errmessage{(Inkscape) Transparency is used (non-zero) for the text in Inkscape, but the package 'transparent.sty' is not loaded}
    \renewcommand\transparent[1]{}
  }
  \providecommand\rotatebox[2]{#2}
  \newcommand*\fsize{\dimexpr\f@size pt\relax}
  \newcommand*\lineheight[1]{\fontsize{\fsize}{#1\fsize}\selectfont}
  \ifx\svgwidth\undefined
    \setlength{\unitlength}{272.63429657bp}
    \ifx\svgscale\undefined
      \relax
    \else
      \setlength{\unitlength}{\unitlength * \real{\svgscale}}
    \fi
  \else
    \setlength{\unitlength}{\svgwidth}
  \fi
  \global\let\svgwidth\undefined
  \global\let\svgscale\undefined
  \makeatother
  \begin{picture}(1,0.2640913)
    \lineheight{1}
    \setlength\tabcolsep{0pt}
    \put(0.00190036,0.2474225){\color[rgb]{0,0,0}\makebox(0,0)[lt]{\lineheight{1.25}\smash{\begin{tabular}[t]{l}$G':$\end{tabular}}}}
    \put(0.33625675,0.24686522){\color[rgb]{0,0,0}\makebox(0,0)[lt]{\lineheight{1.25}\smash{\begin{tabular}[t]{l}$T^2(G',v'):$\end{tabular}}}}
    \put(0,0){\includegraphics[width=\unitlength,page=1]{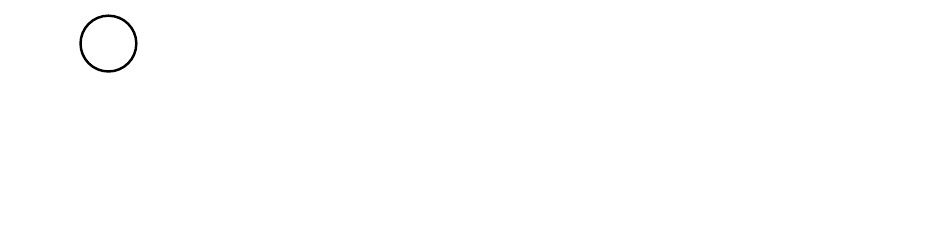}}
    \put(0.1066434,0.20731289){\color[rgb]{0,0,0}\makebox(0,0)[lt]{\lineheight{1.25}\smash{\begin{tabular}[t]{l}$1$\end{tabular}}}}
    \put(0,0){\includegraphics[width=\unitlength,page=2]{unfolding_example_b.pdf}}
    \put(0.02288579,0.15521926){\color[rgb]{0,0,0}\makebox(0,0)[lt]{\lineheight{1.25}\smash{\begin{tabular}[t]{l}$2$\end{tabular}}}}
    \put(0,0){\includegraphics[width=\unitlength,page=3]{unfolding_example_b.pdf}}
    \put(0.1919702,0.15509635){\color[rgb]{0,0,0}\makebox(0,0)[lt]{\lineheight{1.25}\smash{\begin{tabular}[t]{l}$3$\end{tabular}}}}
    \put(0,0){\includegraphics[width=\unitlength,page=4]{unfolding_example_b.pdf}}
    \put(0.0229097,0.07259358){\color[rgb]{0,0,0}\makebox(0,0)[lt]{\lineheight{1.25}\smash{\begin{tabular}[t]{l}$4$\end{tabular}}}}
    \put(0,0){\includegraphics[width=\unitlength,page=5]{unfolding_example_b.pdf}}
    \put(0.19199401,0.07247069){\color[rgb]{0,0,0}\makebox(0,0)[lt]{\lineheight{1.25}\smash{\begin{tabular}[t]{l}$5$\end{tabular}}}}
    \put(0,0){\includegraphics[width=\unitlength,page=6]{unfolding_example_b.pdf}}
    \put(0.10670121,0.02025302){\color[rgb]{0,0,0}\makebox(0,0)[lt]{\lineheight{1.25}\smash{\begin{tabular}[t]{l}$6$\end{tabular}}}}
    \put(0,0){\includegraphics[width=\unitlength,page=7]{unfolding_example_b.pdf}}
    \put(0.37780437,0.1138042){\color[rgb]{0,0,0}\makebox(0,0)[lt]{\lineheight{1.25}\smash{\begin{tabular}[t]{l}$2$\end{tabular}}}}
    \put(0,0){\includegraphics[width=\unitlength,page=8]{unfolding_example_b.pdf}}
    \put(0.34629211,0.0201573){\color[rgb]{0,0,0}\makebox(0,0)[lt]{\lineheight{1.25}\smash{\begin{tabular}[t]{l}$1$\end{tabular}}}}
    \put(0,0){\includegraphics[width=\unitlength,page=9]{unfolding_example_b.pdf}}
    \put(0.41219173,0.02003468){\color[rgb]{0,0,0}\makebox(0,0)[lt]{\lineheight{1.25}\smash{\begin{tabular}[t]{l}$4$\end{tabular}}}}
    \put(0,0){\includegraphics[width=\unitlength,page=10]{unfolding_example_b.pdf}}
    \put(0.51625536,0.1138042){\color[rgb]{0,0,0}\makebox(0,0)[lt]{\lineheight{1.25}\smash{\begin{tabular}[t]{l}$3$\end{tabular}}}}
    \put(0,0){\includegraphics[width=\unitlength,page=11]{unfolding_example_b.pdf}}
    \put(0.48474311,0.0201573){\color[rgb]{0,0,0}\makebox(0,0)[lt]{\lineheight{1.25}\smash{\begin{tabular}[t]{l}$1$\end{tabular}}}}
    \put(0,0){\includegraphics[width=\unitlength,page=12]{unfolding_example_b.pdf}}
    \put(0.55064272,0.02003468){\color[rgb]{0,0,0}\makebox(0,0)[lt]{\lineheight{1.25}\smash{\begin{tabular}[t]{l}$5$\end{tabular}}}}
    \put(0,0){\includegraphics[width=\unitlength,page=13]{unfolding_example_b.pdf}}
    \put(0.68827409,0.1137105){\color[rgb]{0,0,0}\makebox(0,0)[lt]{\lineheight{1.25}\smash{\begin{tabular}[t]{l}$4$\end{tabular}}}}
    \put(0,0){\includegraphics[width=\unitlength,page=14]{unfolding_example_b.pdf}}
    \put(0.62272298,0.02019834){\color[rgb]{0,0,0}\makebox(0,0)[lt]{\lineheight{1.25}\smash{\begin{tabular}[t]{l}$1$\end{tabular}}}}
    \put(0,0){\includegraphics[width=\unitlength,page=15]{unfolding_example_b.pdf}}
    \put(0.6886226,0.02007568){\color[rgb]{0,0,0}\makebox(0,0)[lt]{\lineheight{1.25}\smash{\begin{tabular}[t]{l}$2$\end{tabular}}}}
    \put(0,0){\includegraphics[width=\unitlength,page=16]{unfolding_example_b.pdf}}
    \put(0.75458658,0.02003468){\color[rgb]{0,0,0}\makebox(0,0)[lt]{\lineheight{1.25}\smash{\begin{tabular}[t]{l}$6$\end{tabular}}}}
    \put(0,0){\includegraphics[width=\unitlength,page=17]{unfolding_example_b.pdf}}
    \put(0.89361194,0.11371041){\color[rgb]{0,0,0}\makebox(0,0)[lt]{\lineheight{1.25}\smash{\begin{tabular}[t]{l}$6$\end{tabular}}}}
    \put(0,0){\includegraphics[width=\unitlength,page=18]{unfolding_example_b.pdf}}
    \put(0.82806085,0.02019834){\color[rgb]{0,0,0}\makebox(0,0)[lt]{\lineheight{1.25}\smash{\begin{tabular}[t]{l}$1$\end{tabular}}}}
    \put(0,0){\includegraphics[width=\unitlength,page=19]{unfolding_example_b.pdf}}
    \put(0.8939605,0.02007567){\color[rgb]{0,0,0}\makebox(0,0)[lt]{\lineheight{1.25}\smash{\begin{tabular}[t]{l}$4$\end{tabular}}}}
    \put(0,0){\includegraphics[width=\unitlength,page=20]{unfolding_example_b.pdf}}
    \put(0.95992445,0.02003468){\color[rgb]{0,0,0}\makebox(0,0)[lt]{\lineheight{1.25}\smash{\begin{tabular}[t]{l}$5$\end{tabular}}}}
    \put(0,0){\includegraphics[width=\unitlength,page=21]{unfolding_example_b.pdf}}
    \put(0.62625678,0.20718508){\color[rgb]{0,0,0}\makebox(0,0)[lt]{\lineheight{1.25}\smash{\begin{tabular}[t]{l}$1$\end{tabular}}}}
    \put(0,0){\includegraphics[width=\unitlength,page=22]{unfolding_example_b.pdf}}
    \put(0.13890385,0.24112095){\color[rgb]{0,0,0}\makebox(0,0)[lt]{\lineheight{1.25}\smash{\begin{tabular}[t]{l}$v'$\end{tabular}}}}
  \end{picture}
\endgroup
 }
			\caption{}
			\label{fig:unfolding_example_b}
		\end{subfigure}
		\caption{
			Unfolding trees $T^2(G,v)$ and $T^2(G',v')$. 
			As $v$ and $v'$ have structurally similar roles in $G$, resp. $G'$, their unfolding trees differ only slightly (labeled yellow). The vertex corresponding to $v$, resp. $v'$, appears again several times at depth $2$ of $T^2(G,v)$, resp. $T^2(G',v')$. }
		\label{fig:unfolding_example}
	\end{figure}

	Another view of the Weisfeiler-Lehman procedure is that for each iteration $i$, it implicitly constructs tree patterns of depth $i$ which are being compressed into labels.
	Each such tree, denoted $T^i(G,v)$, is called the depth-$i$ \emph{unfolding tree} (or simply, $i$-unfolding tree) of $G$ at $v$. 
	Figure~\ref{fig:unfolding_example} visualizes this concept and illustrates that there is a function from the vertices in the unfolding tree of $G$ at $v$ into the corresponding vertices of graph $G$.
	Thus, a node of $G$ can appear several times in $T^i(G,v)$ for $i >1$.
	It is easy to see that there is a bijection between labels in $\Sigma_i$ and the set of (pairwise non-isomorphic) $i$-unfolding trees. 
	
	\subsection{The Structure and Depth Preserving Tree Edit Distance}
	While the {\em strict} comparison of labels, or equivalently, that of unfolding trees is advantageous for the original intention of the Weisfeiler-Lehman method, it is a severe drawback of all Weisfeiler-Lehman graph kernels, including the Weisfeiler-Lehman subtree kernel.
	The reason is that comparing unfolding trees with each other by {\em equality} (i.e., tree isomorphism), or equivalently, taking merely into account whether the labels of vertices and those of their neighborhoods differ or not,
	is too restrictive, as in case of kernels, we are interested in defining similarities. 
	Our typical observation is that the $i$-unfolding trees (i.e., labels at iteration $i$) of most vertices will be unique for very small values of $i$.
	In other words, the limitation of the Weisfeiler-Lehman graph kernels is that two structurally completely different unfolding trees are treated identically to two unfolding trees which differ by only very little.
	
	To overcome this drawback, we propose a finer comparison by defining a new {\em similarity measure} between unfolding trees that employs a specialized form of the well-known {\em tree edit distance}.
	On an abstract level, the tree edit distance measures the minimum amount of edit operations necessary to turn one tree into another.
	Calculating this distance is NP-hard in general (see, e.g., \cite{billeSurveyTreeEdit2005}). 
	However, for our purpose it suffices to consider a constrained tree edit distance which preserves essential properties of unfolding trees. 
	Below we show that, in contrast to the general case, this variant can be calculated efficiently.
	
	The construction procedure of unfolding trees as demonstrated above shows that they reflect the neighborhoods of a specific vertex.
	Therefore, we require the edit scripts between unfolding trees to preserve the neighborhood relationships of vertex pairs as well as the depth of vertices. 
	This leads to the following definition of constrained tree edit scripts (cf. \cite{billeSurveyTreeEdit2005}):  

	\begin{definition}
		\label{def:1}
		A \emph{structure and depth preserving mapping} $(\sdm)$ between two rooted trees $T$ and $T'$ is a triple $(M,T,T')$ with $M \subseteq V(T) \times V(T')$ satisfying
		\begin{enumerate}
			\item $\forall (v_1,v'_1),(v_2,v_2') \in M: v_1 = v_2 \Leftrightarrow v'_1 = v'_2$, \\ \hspace*{\fill} (definite)
			\item $(r(T),r(T')) \in M$, \hspace*{\fill} (root preserving)
			\item $\forall (v,v') \in M: (par(v),par(v')) \in M$. \\ \hspace*{\fill} (structure preserving) \\
		\end{enumerate}
		The set of all structure and depth preserving mappings between $T$ and $T'$ is denoted by $\sdm(T,T')$.
	\end{definition}
	
	$\sdm$s represent sequences of edit operations subject to the above constraints that transform trees into trees.  
	More precisely, for an $\sdm$ $(M,T,T')$ let $T=T_0,T_1,\ldots,T_k$ be a sequence of trees such that $T_{i+1}$ is obtained from $T_i$ by applying one of the following atomic transformations:
	
	\begin{description}
		\item[relabel:] ~~If $(v,v') \in M$, then replace the label of $v$ in $T_i$ by that of $v'$.
		\item[delete:] ~~If $v$ is a leaf in $T_i$ and it does not occur in a pair of $M$, then remove $v$ from $T_i$. 
		\item[insert:] ~~If $v'$  is a vertex in $T'$ which does not occur in a pair of $M$ and for which the corresponding parent $u$ already exists in $T_i$, then add a child to $u$ with the label of $v'$.
		\end{description}
	The proof of the following claim is straightforward.
	\begin{proposition}
		Let $(M,T,T')$ be an $\sdm$ and $T_0 = T,T_1,\ldots,T_k$ be a sequence of trees obtained by the above atomic transformations such that every $v \in T$ and $v' \in T'$ has been considered in exactly one transformation. Then $T_k = T'$.
	\end{proposition}
	
	Note that $\sdm$s uphold some essential properties of unfolding trees.
	In particular, they ensure that siblings are preserved (i.e., for any $\sdm$ $(M,T,T')$, $v_1'$ and $v_2'$ are siblings in $T'$ whenever $(v_1,v_1'),(v_2,v_2') \in M$ and $v_1,v_2$ are siblings in $T$) and that vertices can only be mapped onto vertices of the same depth.
	Recall, that our goal is to measure similarities between neighborhoods of vertices. 
	It is thus essential that roots are being preserved; this is guaranteed by the second constraint in Def.~\ref{def:1}.                           
	Furthermore, Def.~\ref{def:1} implies that $M$ maps a connected subtree of $T$ onto a connected subtree of $T'$.
	That is, the first (resp. second) components of the pairs in $M$ form a connected subtree of $T$ (resp. $T'$). 
	
\begin{figure}[t]
		\centering
		\begin{subfigure}[b]{0.24\textwidth}
			\resizebox{0.95\textwidth}{!}{\begingroup
  \makeatletter
  \providecommand\color[2][]{
    \errmessage{(Inkscape) Color is used for the text in Inkscape, but the package 'color.sty' is not loaded}
    \renewcommand\color[2][]{}
  }
  \providecommand\transparent[1]{
    \errmessage{(Inkscape) Transparency is used (non-zero) for the text in Inkscape, but the package 'transparent.sty' is not loaded}
    \renewcommand\transparent[1]{}
  }
  \providecommand\rotatebox[2]{#2}
  \newcommand*\fsize{\dimexpr\f@size pt\relax}
  \newcommand*\lineheight[1]{\fontsize{\fsize}{#1\fsize}\selectfont}
  \ifx\svgwidth\undefined
    \setlength{\unitlength}{178.2961135bp}
    \ifx\svgscale\undefined
      \relax
    \else
      \setlength{\unitlength}{\unitlength * \real{\svgscale}}
    \fi
  \else
    \setlength{\unitlength}{\svgwidth}
  \fi
  \global\let\svgwidth\undefined
  \global\let\svgscale\undefined
  \makeatother
  \begin{picture}(1,0.87240441)
    \lineheight{1}
    \setlength\tabcolsep{0pt}
    \put(0.21335098,0.42186999){\color[rgb]{0,0,0}\makebox(0,0)[lt]{\lineheight{1.25}\smash{\begin{tabular}[t]{l}$\downarrow$\end{tabular}}}}
    \put(0.51084177,0.17305425){\color[rgb]{0,0,0}\makebox(0,0)[lt]{\lineheight{1.25}\smash{\begin{tabular}[t]{l}$\rightarrow$\end{tabular}}}}
    \put(0.81127763,0.41519992){\color[rgb]{0,0,0}\makebox(0,0)[lt]{\lineheight{1.25}\smash{\begin{tabular}[t]{l}$\uparrow$\end{tabular}}}}
    \put(0,0){\includegraphics[width=\unitlength,page=1]{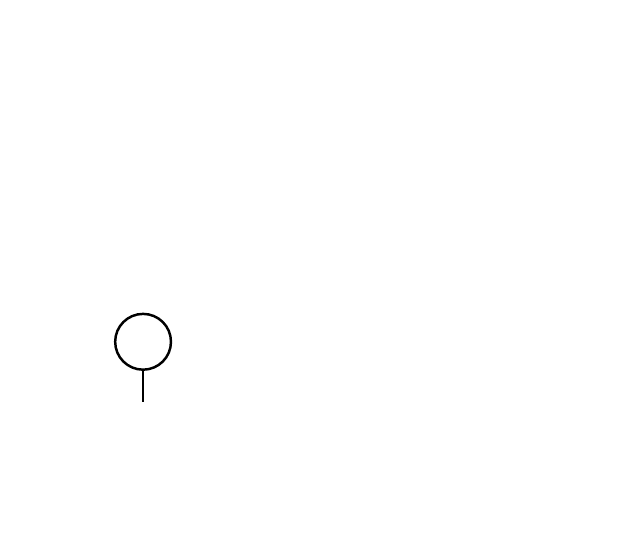}}
    \put(0.21905241,0.30390241){\color[rgb]{0,0,0}\makebox(0,0)[lt]{\lineheight{1.25}\smash{\begin{tabular}[t]{l}$1$\end{tabular}}}}
    \put(0,0){\includegraphics[width=\unitlength,page=2]{structure_preserving_pos_example.pdf}}
    \put(0.21892268,0.16114571){\color[rgb]{0,0,0}\makebox(0,0)[lt]{\lineheight{1.25}\smash{\begin{tabular}[t]{l}$2$\end{tabular}}}}
    \put(0,0){\includegraphics[width=\unitlength,page=3]{structure_preserving_pos_example.pdf}}
    \put(0.14973912,0.0310107){\color[rgb]{0,0,0}\makebox(0,0)[lt]{\lineheight{1.25}\smash{\begin{tabular}[t]{l}$1$\end{tabular}}}}
    \put(0,0){\includegraphics[width=\unitlength,page=4]{structure_preserving_pos_example.pdf}}
    \put(0.29313503,0.03063518){\color[rgb]{0,0,0}\makebox(0,0)[lt]{\lineheight{1.25}\smash{\begin{tabular}[t]{l}$3$\end{tabular}}}}
    \put(0,0){\includegraphics[width=\unitlength,page=5]{structure_preserving_pos_example.pdf}}
    \put(0.81637491,0.30469113){\color[rgb]{0,0,0}\makebox(0,0)[lt]{\lineheight{1.25}\smash{\begin{tabular}[t]{l}$1$\end{tabular}}}}
    \put(0,0){\includegraphics[width=\unitlength,page=6]{structure_preserving_pos_example.pdf}}
    \put(0.81522909,0.16973268){\color[rgb]{0,0,0}\makebox(0,0)[lt]{\lineheight{1.25}\smash{\begin{tabular}[t]{l}$2$\end{tabular}}}}
    \put(0,0){\includegraphics[width=\unitlength,page=7]{structure_preserving_pos_example.pdf}}
    \put(0.81587414,0.03585886){\color[rgb]{0,0,0}\makebox(0,0)[lt]{\lineheight{1.25}\smash{\begin{tabular}[t]{l}$2$\end{tabular}}}}
    \put(0,0){\includegraphics[width=\unitlength,page=8]{structure_preserving_pos_example.pdf}}
    \put(0.93403102,0.03548357){\color[rgb]{0,0,0}\makebox(0,0)[lt]{\lineheight{1.25}\smash{\begin{tabular}[t]{l}$3$\end{tabular}}}}
    \put(0,0){\includegraphics[width=\unitlength,page=9]{structure_preserving_pos_example.pdf}}
    \put(0.69780996,0.03569743){\color[rgb]{0,0,0}\makebox(0,0)[lt]{\lineheight{1.25}\smash{\begin{tabular}[t]{l}$1$\end{tabular}}}}
    \put(0,0){\includegraphics[width=\unitlength,page=10]{structure_preserving_pos_example.pdf}}
    \put(0.81394073,0.79502261){\color[rgb]{0,0,0}\makebox(0,0)[lt]{\lineheight{1.25}\smash{\begin{tabular}[t]{l}$1$\end{tabular}}}}
    \put(0,0){\includegraphics[width=\unitlength,page=11]{structure_preserving_pos_example.pdf}}
    \put(0.81279431,0.66006431){\color[rgb]{0,0,0}\makebox(0,0)[lt]{\lineheight{1.25}\smash{\begin{tabular}[t]{l}$1$\end{tabular}}}}
    \put(0,0){\includegraphics[width=\unitlength,page=12]{structure_preserving_pos_example.pdf}}
    \put(0.81343928,0.52619037){\color[rgb]{0,0,0}\makebox(0,0)[lt]{\lineheight{1.25}\smash{\begin{tabular}[t]{l}$2$\end{tabular}}}}
    \put(0,0){\includegraphics[width=\unitlength,page=13]{structure_preserving_pos_example.pdf}}
    \put(0.93159659,0.52581485){\color[rgb]{0,0,0}\makebox(0,0)[lt]{\lineheight{1.25}\smash{\begin{tabular}[t]{l}$3$\end{tabular}}}}
    \put(0,0){\includegraphics[width=\unitlength,page=14]{structure_preserving_pos_example.pdf}}
    \put(0.69537556,0.52602868){\color[rgb]{0,0,0}\makebox(0,0)[lt]{\lineheight{1.25}\smash{\begin{tabular}[t]{l}$1$\end{tabular}}}}
    \put(0,0){\includegraphics[width=\unitlength,page=15]{structure_preserving_pos_example.pdf}}
    \put(0.21567862,0.79679277){\color[rgb]{0,0,0}\makebox(0,0)[lt]{\lineheight{1.25}\smash{\begin{tabular}[t]{l}$1$\end{tabular}}}}
    \put(0,0){\includegraphics[width=\unitlength,page=16]{structure_preserving_pos_example.pdf}}
    \put(0.09489026,0.65983846){\color[rgb]{0,0,0}\makebox(0,0)[lt]{\lineheight{1.25}\smash{\begin{tabular}[t]{l}$2$\end{tabular}}}}
    \put(0,0){\includegraphics[width=\unitlength,page=17]{structure_preserving_pos_example.pdf}}
    \put(0.3392423,0.65946269){\color[rgb]{0,0,0}\makebox(0,0)[lt]{\lineheight{1.25}\smash{\begin{tabular}[t]{l}$3$\end{tabular}}}}
    \put(0,0){\includegraphics[width=\unitlength,page=18]{structure_preserving_pos_example.pdf}}
    \put(0.3409732,0.52456949){\color[rgb]{0,0,0}\makebox(0,0)[lt]{\lineheight{1.25}\smash{\begin{tabular}[t]{l}$1$\end{tabular}}}}
    \put(0,0){\includegraphics[width=\unitlength,page=19]{structure_preserving_pos_example.pdf}}
    \put(0.15305845,0.52438955){\color[rgb]{0,0,0}\makebox(0,0)[lt]{\lineheight{1.25}\smash{\begin{tabular}[t]{l}$3$\end{tabular}}}}
    \put(0,0){\includegraphics[width=\unitlength,page=20]{structure_preserving_pos_example.pdf}}
    \put(0.03499487,0.52422786){\color[rgb]{0,0,0}\makebox(0,0)[lt]{\lineheight{1.25}\smash{\begin{tabular}[t]{l}$1$\end{tabular}}}}
    \put(0,0){\includegraphics[width=\unitlength,page=21]{structure_preserving_pos_example.pdf}}
  \end{picture}
\endgroup
 }
			\caption{}
			\label{fig:sdm_example_pos}
		\end{subfigure}
		\hfill
		\begin{subfigure}[b]{0.24\textwidth}
			\resizebox{0.95\textwidth}{!}{\begingroup
  \makeatletter
  \providecommand\color[2][]{
    \errmessage{(Inkscape) Color is used for the text in Inkscape, but the package 'color.sty' is not loaded}
    \renewcommand\color[2][]{}
  }
  \providecommand\transparent[1]{
    \errmessage{(Inkscape) Transparency is used (non-zero) for the text in Inkscape, but the package 'transparent.sty' is not loaded}
    \renewcommand\transparent[1]{}
  }
  \providecommand\rotatebox[2]{#2}
  \newcommand*\fsize{\dimexpr\f@size pt\relax}
  \newcommand*\lineheight[1]{\fontsize{\fsize}{#1\fsize}\selectfont}
  \ifx\svgwidth\undefined
    \setlength{\unitlength}{178.2961135bp}
    \ifx\svgscale\undefined
      \relax
    \else
      \setlength{\unitlength}{\unitlength * \real{\svgscale}}
    \fi
  \else
    \setlength{\unitlength}{\svgwidth}
  \fi
  \global\let\svgwidth\undefined
  \global\let\svgscale\undefined
  \makeatother
  \begin{picture}(1,0.87240441)
    \lineheight{1}
    \setlength\tabcolsep{0pt}
    \put(0,0){\includegraphics[width=\unitlength,page=1]{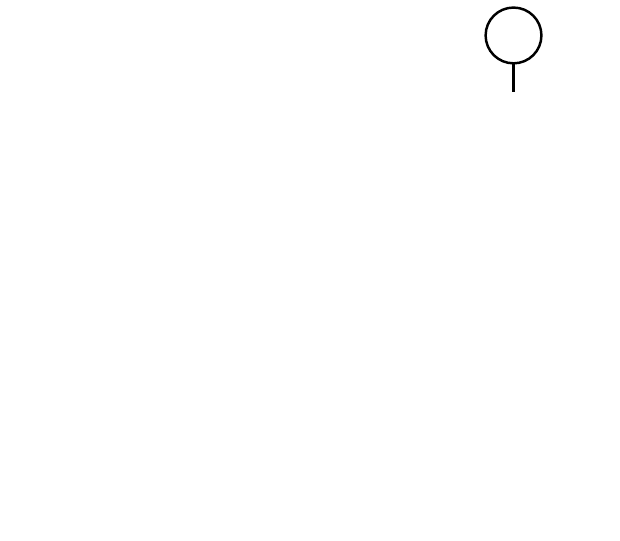}}
    \put(0.8174397,0.79873408){\color[rgb]{0,0,0}\makebox(0,0)[lt]{\lineheight{1.25}\smash{\begin{tabular}[t]{l}$1$\end{tabular}}}}
    \put(0,0){\includegraphics[width=\unitlength,page=2]{structure_preserving_neg_example.pdf}}
    \put(0.81629331,0.66377559){\color[rgb]{0,0,0}\makebox(0,0)[lt]{\lineheight{1.25}\smash{\begin{tabular}[t]{l}$1$\end{tabular}}}}
    \put(0,0){\includegraphics[width=\unitlength,page=3]{structure_preserving_neg_example.pdf}}
    \put(0.81693827,0.52990167){\color[rgb]{0,0,0}\makebox(0,0)[lt]{\lineheight{1.25}\smash{\begin{tabular}[t]{l}$2$\end{tabular}}}}
    \put(0,0){\includegraphics[width=\unitlength,page=4]{structure_preserving_neg_example.pdf}}
    \put(0.93509556,0.5295262){\color[rgb]{0,0,0}\makebox(0,0)[lt]{\lineheight{1.25}\smash{\begin{tabular}[t]{l}$3$\end{tabular}}}}
    \put(0,0){\includegraphics[width=\unitlength,page=5]{structure_preserving_neg_example.pdf}}
    \put(0.69887452,0.52974004){\color[rgb]{0,0,0}\makebox(0,0)[lt]{\lineheight{1.25}\smash{\begin{tabular}[t]{l}$1$\end{tabular}}}}
    \put(0,0){\includegraphics[width=\unitlength,page=6]{structure_preserving_neg_example.pdf}}
    \put(0.21917756,0.80050423){\color[rgb]{0,0,0}\makebox(0,0)[lt]{\lineheight{1.25}\smash{\begin{tabular}[t]{l}$1$\end{tabular}}}}
    \put(0,0){\includegraphics[width=\unitlength,page=7]{structure_preserving_neg_example.pdf}}
    \put(0.09838929,0.66354964){\color[rgb]{0,0,0}\makebox(0,0)[lt]{\lineheight{1.25}\smash{\begin{tabular}[t]{l}$2$\end{tabular}}}}
    \put(0,0){\includegraphics[width=\unitlength,page=8]{structure_preserving_neg_example.pdf}}
    \put(0.34274133,0.66317386){\color[rgb]{0,0,0}\makebox(0,0)[lt]{\lineheight{1.25}\smash{\begin{tabular}[t]{l}$3$\end{tabular}}}}
    \put(0,0){\includegraphics[width=\unitlength,page=9]{structure_preserving_neg_example.pdf}}
    \put(0.34447211,0.52828087){\color[rgb]{0,0,0}\makebox(0,0)[lt]{\lineheight{1.25}\smash{\begin{tabular}[t]{l}$1$\end{tabular}}}}
    \put(0,0){\includegraphics[width=\unitlength,page=10]{structure_preserving_neg_example.pdf}}
    \put(0.1565574,0.5281009){\color[rgb]{0,0,0}\makebox(0,0)[lt]{\lineheight{1.25}\smash{\begin{tabular}[t]{l}$3$\end{tabular}}}}
    \put(0,0){\includegraphics[width=\unitlength,page=11]{structure_preserving_neg_example.pdf}}
    \put(0.0384938,0.52793921){\color[rgb]{0,0,0}\makebox(0,0)[lt]{\lineheight{1.25}\smash{\begin{tabular}[t]{l}$1$\end{tabular}}}}
    \put(0,0){\includegraphics[width=\unitlength,page=12]{structure_preserving_neg_example.pdf}}
    \put(0.21685001,0.36669044){\color[rgb]{0,0,0}\makebox(0,0)[lt]{\lineheight{1.25}\smash{\begin{tabular}[t]{l}$\downarrow$\end{tabular}}}}
    \put(0,0){\includegraphics[width=\unitlength,page=13]{structure_preserving_neg_example.pdf}}
    \put(0.21917756,0.16953075){\color[rgb]{0,0,0}\makebox(0,0)[lt]{\lineheight{1.25}\smash{\begin{tabular}[t]{l}$1$\end{tabular}}}}
    \put(0,0){\includegraphics[width=\unitlength,page=14]{structure_preserving_neg_example.pdf}}
    \put(0.81450228,0.30579589){\color[rgb]{0,0,0}\makebox(0,0)[lt]{\lineheight{1.25}\smash{\begin{tabular}[t]{l}$1$\end{tabular}}}}
    \put(0,0){\includegraphics[width=\unitlength,page=15]{structure_preserving_neg_example.pdf}}
    \put(0.81337329,0.16861579){\color[rgb]{0,0,0}\makebox(0,0)[lt]{\lineheight{1.25}\smash{\begin{tabular}[t]{l}$1$\end{tabular}}}}
    \put(0,0){\includegraphics[width=\unitlength,page=16]{structure_preserving_neg_example.pdf}}
    \put(0.51434055,0.17676567){\color[rgb]{0,0,0}\makebox(0,0)[lt]{\lineheight{1.25}\smash{\begin{tabular}[t]{l}$\rightarrow$\end{tabular}}}}
    \put(0.80636343,0.41049848){\color[rgb]{0,0,0}\makebox(0,0)[lt]{\lineheight{1.25}\smash{\begin{tabular}[t]{l}$\uparrow$\end{tabular}}}}
    \put(0,0){\includegraphics[width=\unitlength,page=17]{structure_preserving_neg_example.pdf}}
    \put(0.21972424,0.03469126){\color[rgb]{0,0,0}\makebox(0,0)[lt]{\lineheight{1.25}\smash{\begin{tabular}[t]{l}$1$\end{tabular}}}}
    \put(0,0){\includegraphics[width=\unitlength,page=18]{structure_preserving_neg_example.pdf}}
    \put(0.33788139,0.03431569){\color[rgb]{0,0,0}\makebox(0,0)[lt]{\lineheight{1.25}\smash{\begin{tabular}[t]{l}$3$\end{tabular}}}}
    \put(0,0){\includegraphics[width=\unitlength,page=19]{structure_preserving_neg_example.pdf}}
    \put(0.1016604,0.03452936){\color[rgb]{0,0,0}\makebox(0,0)[lt]{\lineheight{1.25}\smash{\begin{tabular}[t]{l}$1$\end{tabular}}}}
    \put(0,0){\includegraphics[width=\unitlength,page=20]{structure_preserving_neg_example.pdf}}
    \put(0.81339861,0.03537295){\color[rgb]{0,0,0}\makebox(0,0)[lt]{\lineheight{1.25}\smash{\begin{tabular}[t]{l}$1$\end{tabular}}}}
    \put(0,0){\includegraphics[width=\unitlength,page=21]{structure_preserving_neg_example.pdf}}
    \put(0.93155572,0.03499726){\color[rgb]{0,0,0}\makebox(0,0)[lt]{\lineheight{1.25}\smash{\begin{tabular}[t]{l}$3$\end{tabular}}}}
    \put(0,0){\includegraphics[width=\unitlength,page=22]{structure_preserving_neg_example.pdf}}
    \put(0.69533477,0.03521109){\color[rgb]{0,0,0}\makebox(0,0)[lt]{\lineheight{1.25}\smash{\begin{tabular}[t]{l}$1$\end{tabular}}}}
    \put(0,0){\includegraphics[width=\unitlength,page=23]{structure_preserving_neg_example.pdf}}
  \end{picture}
\endgroup
 }
			\caption{}
			\label{fig:sdm_example_neg}
		\end{subfigure}
		\caption{Two mappings from one unfolding tree into another. \ref{fig:sdm_example_pos} depicts a mapping which is structure and depth preserving whereas \ref{fig:sdm_example_neg} is not. Dashed lines correspond to pairs contained in the respective mappings $M$, red vertices are being deleted, blue vertices inserted and yellow vertices relabeled. }
		\label{fig:sdm_example}
	\end{figure}

	Figure \ref{fig:sdm_example} demonstrates the motivation of $\sdm$s. 
	The mapping displayed in (a) is a structure and depth preserving mapping from $T$ into $T'$ which visibly preserves the depth as well as pairwise sibling relationships of all mapped vertices.
	In contrast, while the edit script in (b) is valid for more general definitions of edit operation sequences, the transformation constructs a tree which heavily distorts neighborhood relationships and arbitrarily inserts nodes such that the set of vertices in $T'$ touched by a line preserve only very little of the topology of those in $T$. 
	In particular, leafs that have distance $4$ from each other in $T$ are mapped onto vertices in $T'$ which are now direct siblings.
	Furthermore, the mapping does not maintain root nodes, as a root is mapped to a non-root node. 
	
	Using these notions, we define the distance between two unfolding trees.

	\begin{definition}
		\label{def:2}
		Let $T,T'$ be unfolding trees over the vertex label alphabet $\Sigma$ and $\gamma: \Sigma^\bot \times \Sigma^\bot \to \mathbb{R}$ a cost function (i.e., metric), where $\bot$ is the blank symbol. Then the {\em cost} $\gamma(M)$ for an $\sdm$ $(M,T,T')$ is 
		\begin{gather*}
			\gamma(M)= \\ \sum_{(v,v') \in M}\gamma(\ell(v),\ell(v')) + \sum_{v \in N}\gamma(\ell(v),\bot) + \sum_{v' \in N'}\gamma(\bot, \ell(v')) 
		\end{gather*}
		where $N$ (resp. $N'$) are the vertices of $T$ (resp. $T'$) that do not occur in any pair of $M$.
		The \emph{structure and depth preserving tree edit distance} from $T$ into $T'$, denoted $\sdted(T,T')$, is then defined by 
		\[ \sdted(T,T') = \min\{\gamma(M): (M,T,T') \in \sdm(T,T') \} \] 
	\end{definition}
	
	Thus, the cost of an $\sdm$ $(M,T,T')$ is defined by the sum of the individual costs of relabeling, insertion, and deletion operations over all vertices of $T$ and $T'$, where the cost of the insertion (resp. deletion) of a vertex $v$ is given by $\gamma(\ell(v),\bot)$ (resp. $\gamma(\bot,\ell(v))$).
	The structure and depth preserving tree edit distance between trees $T$ and $T'$ is then simply the minimal cost over all possible mappings.

	\begin{algorithm}[t]
		\caption{\sc Compute SdTed}
		\label{alg:1}
		\begin{tabbing}
			\textbf{input:} Trees $T,T'$, cost function $\gamma:\Sigma^{\bot} \times \Sigma^{\bot} \rightarrow \bbR$ \\ 
			\textbf{output:} Structure and depth preserving tree edit \\ distance between $T$ and $T'$
		\end{tabbing}
		{\sc SdTed}$(T,T')$:
		\begin{algorithmic}[1]
			\State $F := F(r(T))$, $F' := F(r(T'))$ 
			\State Pad $F$ and $F'$ with empty trees $T_\bot$ such that $|F| = |F'| = deg(r(T))+deg(r(T'))$ 
			\ForAll{$T_i \in F, ~T'_j \in F'$}
				\[
				\delta_{ij} =
				\begin{cases}
				\sdted(T_i,T'_j) & \text{if $T_i \not \equiv T_\bot$ and $T'_j \not \equiv T_\bot$} \\
				\sum\limits_{v\in V(T_i)} \gamma(\ell(v),\bot) & \text{if $T_i \not \equiv T_\bot$ and $T'_j \equiv T_\bot$} \\
				\sum\limits_{v'\in V(T'_j)} \gamma(\ell(v'),\bot) & \text{if $T_i \equiv T_\bot$ and $T'_j \not \equiv T_\bot$} \\
				0 & \text{o/w .}
				\end{cases} 
				\]
			\EndFor
			\State Let $S \subseteq F \times F'$ be a minimum cost perfect bipartite matching w.r.t. distances $\delta$
			\State \Return $\gamma(\ell(r(T)),\ell(r(T'))) + \sum_{(T_i,T_j) \in S} \delta_{ij}$ 
		\end{algorithmic}
	\end{algorithm}

	\subsection{The Unfolding Tree Edit Distance Algorithm}
	\label{ss:tedalgorithm}

	We now show that for any pair of unfolding trees $T,T'$, $\sdted(T,T')$ can efficiently be calculated in a recursive manner.
	It follows from the properties of $\sdm$s that subtrees of $T$ are mapped onto subtrees of $T'$.
	Thus, finding an optimal $\sdm$ (i.e. an $\sdm$ of minimal cost) from $T$ into $T'$ is equivalent to finding the set of optimal $\sdm$s turning the trees below the root of $T$ (i.e. $F(r(T))$) into the trees below the root of $T'$ (i.e., $F(r(T'))$). 
	In order to find this set of optimal $\sdm$s, we need the pairwise distances $\sdted(T_i,T'_j)$ as well as the costs of deleting, resp. inserting, trees $T_i$, resp. $T'_j$.
	The computation of these costs is done in line 3 of Alg.~\ref{alg:1}.
	The first case recursively calculates the $\sdted(T_i,T'_j)$ for all pairs of trees in $F(r(T))$ and $F(r(T'))$.
	The second case considers the instance where the root of some tree $T_i$ is not part of a mapping, which implies that all vertices in $T_i$ are deleted. 
	A similar argument follows for the insertion of trees $T'_j$ (third case of line 3).
	The task of finding an optimal $\sdm$ can in fact be reduced to the {\em minimum cost perfect bipartite matching} problem, as follows:
	Let the sets of trees below the roots of $T$ and $T'$ be $F = \{T_1,\ldots,T_k\}$ and $F' =\{T'_1,\ldots,T'_{k'}\}$, respectively.
	We first expand the set of trees $F$ by $k'$, resp. $F'$ by $k$, auxiliary empty graphs $T_\bot$ (line 2) such that both sets have equal cardinality.
	The distance (c.f. $\delta$ in Alg.~\ref{alg:1}) between a tree and an empty graph is defined as the cost of deleting, resp. inserting that tree. 
	Furthermore, two empty graphs clearly have distance $0$. 
	One can check that the optimal set of $\sdm$s directly corresponds to a perfect bipartite matching of minimum cost between trees in the expanded sets $F$ and $F'$ (line 4) with distances as defined above. 
	Finally, the $\sdted$ between trees $T$ and $T'$ is the cumulative cost of the distance between their roots and the minimal cost perfect bipartite matching between the trees below them (line 5). 
	We have the following result:

	\begin{theorem}
		Given unfolding trees $T,T'$ with labels from $\Sigma$ and a cost function $\gamma:\Sigma^{\bot} \times \Sigma^{\bot}\to \mathbb{R}$ over $\Sigma$ and $\bot$, Alg.~\ref{alg:1} returns $\sdted(T,T')$.
	\end{theorem}

\begin{figure}[t]
		\centering
		\begin{subfigure}[b]{0.23\textwidth}
			\resizebox{1.00\textwidth}{!}{\begingroup
  \makeatletter
  \providecommand\color[2][]{
    \errmessage{(Inkscape) Color is used for the text in Inkscape, but the package 'color.sty' is not loaded}
    \renewcommand\color[2][]{}
  }
  \providecommand\transparent[1]{
    \errmessage{(Inkscape) Transparency is used (non-zero) for the text in Inkscape, but the package 'transparent.sty' is not loaded}
    \renewcommand\transparent[1]{}
  }
  \providecommand\rotatebox[2]{#2}
  \newcommand*\fsize{\dimexpr\f@size pt\relax}
  \newcommand*\lineheight[1]{\fontsize{\fsize}{#1\fsize}\selectfont}
  \ifx\svgwidth\undefined
    \setlength{\unitlength}{161.27862645bp}
    \ifx\svgscale\undefined
      \relax
    \else
      \setlength{\unitlength}{\unitlength * \real{\svgscale}}
    \fi
  \else
    \setlength{\unitlength}{\svgwidth}
  \fi
  \global\let\svgwidth\undefined
  \global\let\svgscale\undefined
  \makeatother
  \begin{picture}(1,0.4335012)
    \lineheight{1}
    \setlength\tabcolsep{0pt}
    \put(0,0){\includegraphics[width=\unitlength,page=1]{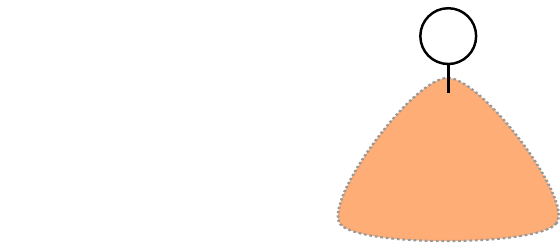}}
    \put(0.78708292,0.35084642){\color[rgb]{0,0,0}\makebox(0,0)[lt]{\lineheight{1.25}\smash{\begin{tabular}[t]{l}$1$\end{tabular}}}}
    \put(0,0){\includegraphics[width=\unitlength,page=2]{ted_calc_trees.pdf}}
    \put(0.78581556,0.20164779){\color[rgb]{0,0,0}\makebox(0,0)[lt]{\lineheight{1.25}\smash{\begin{tabular}[t]{l}$1$\end{tabular}}}}
    \put(0,0){\includegraphics[width=\unitlength,page=3]{ted_calc_trees.pdf}}
    \put(0.78652855,0.05364808){\color[rgb]{0,0,0}\makebox(0,0)[lt]{\lineheight{1.25}\smash{\begin{tabular}[t]{l}$2$\end{tabular}}}}
    \put(0,0){\includegraphics[width=\unitlength,page=4]{ted_calc_trees.pdf}}
    \put(0.91715336,0.05323293){\color[rgb]{0,0,0}\makebox(0,0)[lt]{\lineheight{1.25}\smash{\begin{tabular}[t]{l}$3$\end{tabular}}}}
    \put(0,0){\includegraphics[width=\unitlength,page=5]{ted_calc_trees.pdf}}
    \put(0.65600721,0.05346931){\color[rgb]{0,0,0}\makebox(0,0)[lt]{\lineheight{1.25}\smash{\begin{tabular}[t]{l}$1$\end{tabular}}}}
    \put(0,0){\includegraphics[width=\unitlength,page=6]{ted_calc_trees.pdf}}
    \put(0.25590408,0.35280338){\color[rgb]{0,0,0}\makebox(0,0)[lt]{\lineheight{1.25}\smash{\begin{tabular}[t]{l}$1$\end{tabular}}}}
    \put(0,0){\includegraphics[width=\unitlength,page=7]{ted_calc_trees.pdf}}
    \put(0.12237039,0.20139823){\color[rgb]{0,0,0}\makebox(0,0)[lt]{\lineheight{1.25}\smash{\begin{tabular}[t]{l}$2$\end{tabular}}}}
    \put(0,0){\includegraphics[width=\unitlength,page=8]{ted_calc_trees.pdf}}
    \put(0.39250577,0.2009828){\color[rgb]{0,0,0}\makebox(0,0)[lt]{\lineheight{1.25}\smash{\begin{tabular}[t]{l}$3$\end{tabular}}}}
    \put(0,0){\includegraphics[width=\unitlength,page=9]{ted_calc_trees.pdf}}
    \put(0.3944193,0.05185607){\color[rgb]{0,0,0}\makebox(0,0)[lt]{\lineheight{1.25}\smash{\begin{tabular}[t]{l}$1$\end{tabular}}}}
    \put(0,0){\includegraphics[width=\unitlength,page=10]{ted_calc_trees.pdf}}
    \put(0.18667648,0.05165727){\color[rgb]{0,0,0}\makebox(0,0)[lt]{\lineheight{1.25}\smash{\begin{tabular}[t]{l}$3$\end{tabular}}}}
    \put(0,0){\includegraphics[width=\unitlength,page=11]{ted_calc_trees.pdf}}
    \put(0.05615529,0.0514785){\color[rgb]{0,0,0}\makebox(0,0)[lt]{\lineheight{1.25}\smash{\begin{tabular}[t]{l}$1$\end{tabular}}}}
    \put(0,0){\includegraphics[width=\unitlength,page=12]{ted_calc_trees.pdf}}
    \put(0.02599777,0.40532334){\color[rgb]{0,0,0}\makebox(0,0)[lt]{\lineheight{1.25}\smash{\begin{tabular}[t]{l}$T:$\end{tabular}}}}
    \put(0.60983514,0.40433547){\color[rgb]{0,0,0}\makebox(0,0)[lt]{\lineheight{1.25}\smash{\begin{tabular}[t]{l}$T':$\end{tabular}}}}
    \put(0.19012006,0.18059623){\color[rgb]{0,0,0}\makebox(0,0)[lt]{\lineheight{1.25}\smash{\begin{tabular}[t]{l}$v_2$\end{tabular}}}}
    \put(0.46048461,0.17666591){\color[rgb]{0,0,0}\makebox(0,0)[lt]{\lineheight{1.25}\smash{\begin{tabular}[t]{l}$v_3$\end{tabular}}}}
    \put(0.86077189,0.1892904){\color[rgb]{0,0,0}\makebox(0,0)[lt]{\lineheight{1.25}\smash{\begin{tabular}[t]{l}$v'_2$\end{tabular}}}}
    \put(0.32719193,0.38100927){\color[rgb]{0,0,0}\makebox(0,0)[lt]{\lineheight{1.25}\smash{\begin{tabular}[t]{l}$v_1$\end{tabular}}}}
    \put(0.8594819,0.37161418){\color[rgb]{0,0,0}\makebox(0,0)[lt]{\lineheight{1.25}\smash{\begin{tabular}[t]{l}$v'_1$\end{tabular}}}}
  \end{picture}
\endgroup
 }
			\caption{}
			\label{fig:ted_calc_trees}
		\end{subfigure}
		\hfill
		\begin{subfigure}[b]{0.23\textwidth}
			\resizebox{1.0\textwidth}{!}{\begingroup
  \makeatletter
  \providecommand\color[2][]{
    \errmessage{(Inkscape) Color is used for the text in Inkscape, but the package 'color.sty' is not loaded}
    \renewcommand\color[2][]{}
  }
  \providecommand\transparent[1]{
    \errmessage{(Inkscape) Transparency is used (non-zero) for the text in Inkscape, but the package 'transparent.sty' is not loaded}
    \renewcommand\transparent[1]{}
  }
  \providecommand\rotatebox[2]{#2}
  \newcommand*\fsize{\dimexpr\f@size pt\relax}
  \newcommand*\lineheight[1]{\fontsize{\fsize}{#1\fsize}\selectfont}
  \ifx\svgwidth\undefined
    \setlength{\unitlength}{150.91266902bp}
    \ifx\svgscale\undefined
      \relax
    \else
      \setlength{\unitlength}{\unitlength * \real{\svgscale}}
    \fi
  \else
    \setlength{\unitlength}{\svgwidth}
  \fi
  \global\let\svgwidth\undefined
  \global\let\svgscale\undefined
  \makeatother
  \begin{picture}(1,0.44708821)
    \lineheight{1}
    \setlength\tabcolsep{0pt}
    \put(0.39374359,0.33308283){\color[rgb]{0,0,0}\makebox(0,0)[lt]{\lineheight{1.25}\smash{\begin{tabular}[t]{l}$\bot$\end{tabular}}}}
    \put(0.07619128,0.01554715){\color[rgb]{0,0,0}\makebox(0,0)[lt]{\lineheight{1.25}\smash{\begin{tabular}[t]{l}$\bot$\end{tabular}}}}
    \put(0,0){\includegraphics[width=\unitlength,page=1]{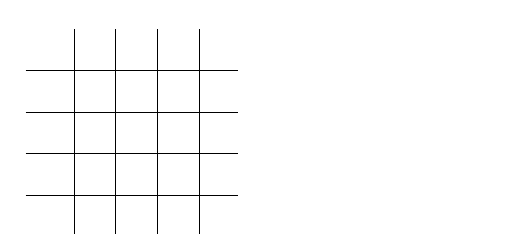}}
    \put(0.16572002,0.25183343){\color[rgb]{0,0,0}\makebox(0,0)[lt]{\lineheight{1.25}\smash{\begin{tabular}[t]{l}$0$\end{tabular}}}}
    \put(0.24621678,0.25124673){\color[rgb]{0,0,0}\makebox(0,0)[lt]{\lineheight{1.25}\smash{\begin{tabular}[t]{l}$1$\end{tabular}}}}
    \put(0.32716782,0.2514038){\color[rgb]{0,0,0}\makebox(0,0)[lt]{\lineheight{1.25}\smash{\begin{tabular}[t]{l}$1$\end{tabular}}}}
    \put(0.40529989,0.25084089){\color[rgb]{0,0,0}\makebox(0,0)[lt]{\lineheight{1.25}\smash{\begin{tabular}[t]{l}$1$\end{tabular}}}}
    \put(0.1654572,0.17185515){\color[rgb]{0,0,0}\makebox(0,0)[lt]{\lineheight{1.25}\smash{\begin{tabular}[t]{l}$1$\end{tabular}}}}
    \put(0.24595424,0.17126845){\color[rgb]{0,0,0}\makebox(0,0)[lt]{\lineheight{1.25}\smash{\begin{tabular}[t]{l}$0$\end{tabular}}}}
    \put(0.32690528,0.17142552){\color[rgb]{0,0,0}\makebox(0,0)[lt]{\lineheight{1.25}\smash{\begin{tabular}[t]{l}$1$\end{tabular}}}}
    \put(0.40503721,0.17086261){\color[rgb]{0,0,0}\makebox(0,0)[lt]{\lineheight{1.25}\smash{\begin{tabular}[t]{l}$1$\end{tabular}}}}
    \put(0.16567918,0.09438013){\color[rgb]{0,0,0}\makebox(0,0)[lt]{\lineheight{1.25}\smash{\begin{tabular}[t]{l}$1$\end{tabular}}}}
    \put(0.24617608,0.09379343){\color[rgb]{0,0,0}\makebox(0,0)[lt]{\lineheight{1.25}\smash{\begin{tabular}[t]{l}$1$\end{tabular}}}}
    \put(0.32712741,0.0939505){\color[rgb]{0,0,0}\makebox(0,0)[lt]{\lineheight{1.25}\smash{\begin{tabular}[t]{l}$0$\end{tabular}}}}
    \put(0.40525905,0.09338788){\color[rgb]{0,0,0}\makebox(0,0)[lt]{\lineheight{1.25}\smash{\begin{tabular}[t]{l}$1$\end{tabular}}}}
    \put(0.16590101,0.01739321){\color[rgb]{0,0,0}\makebox(0,0)[lt]{\lineheight{1.25}\smash{\begin{tabular}[t]{l}$1$\end{tabular}}}}
    \put(0.24639792,0.01680652){\color[rgb]{0,0,0}\makebox(0,0)[lt]{\lineheight{1.25}\smash{\begin{tabular}[t]{l}$1$\end{tabular}}}}
    \put(0.32734924,0.01696329){\color[rgb]{0,0,0}\makebox(0,0)[lt]{\lineheight{1.25}\smash{\begin{tabular}[t]{l}$1$\end{tabular}}}}
    \put(0.40548089,0.01640067){\color[rgb]{0,0,0}\makebox(0,0)[lt]{\lineheight{1.25}\smash{\begin{tabular}[t]{l}$0$\end{tabular}}}}
    \put(-0.00329002,0.41674091){\color[rgb]{0,0,0}\makebox(0,0)[lt]{\lineheight{1.25}\smash{\begin{tabular}[t]{l}$M_r:$\end{tabular}}}}
    \put(0,0){\includegraphics[width=\unitlength,page=2]{ted_calc_matrices.pdf}}
    \put(0.87449546,0.33331613){\color[rgb]{0,0,0}\makebox(0,0)[lt]{\lineheight{1.25}\smash{\begin{tabular}[t]{l}$\bot$\end{tabular}}}}
    \put(0,0){\includegraphics[width=\unitlength,page=3]{ted_calc_matrices.pdf}}
    \put(0.55694057,0.01578045){\color[rgb]{0,0,0}\makebox(0,0)[lt]{\lineheight{1.25}\smash{\begin{tabular}[t]{l}$\bot$\end{tabular}}}}
    \put(0,0){\includegraphics[width=\unitlength,page=4]{ted_calc_matrices.pdf}}
    \put(0.64647118,0.25206673){\color[rgb]{0,0,0}\makebox(0,0)[lt]{\lineheight{1.25}\smash{\begin{tabular}[t]{l}$0$\end{tabular}}}}
    \put(0.7269688,0.25148032){\color[rgb]{0,0,0}\makebox(0,0)[lt]{\lineheight{1.25}\smash{\begin{tabular}[t]{l}$2$\end{tabular}}}}
    \put(0.80791969,0.2516371){\color[rgb]{0,0,0}\makebox(0,0)[lt]{\lineheight{1.25}\smash{\begin{tabular}[t]{l}$2$\end{tabular}}}}
    \put(0.88605162,0.25107448){\color[rgb]{0,0,0}\makebox(0,0)[lt]{\lineheight{1.25}\smash{\begin{tabular}[t]{l}$3$\end{tabular}}}}
    \put(0.64620821,0.17208845){\color[rgb]{0,0,0}\makebox(0,0)[lt]{\lineheight{1.25}\smash{\begin{tabular}[t]{l}$2$\end{tabular}}}}
    \put(0.72670583,0.17150175){\color[rgb]{0,0,0}\makebox(0,0)[lt]{\lineheight{1.25}\smash{\begin{tabular}[t]{l}$0$\end{tabular}}}}
    \put(0.80765716,0.17165882){\color[rgb]{0,0,0}\makebox(0,0)[lt]{\lineheight{1.25}\smash{\begin{tabular}[t]{l}$3$\end{tabular}}}}
    \put(0.88578909,0.1710962){\color[rgb]{0,0,0}\makebox(0,0)[lt]{\lineheight{1.25}\smash{\begin{tabular}[t]{l}$2$\end{tabular}}}}
    \put(0.6464299,0.09461372){\color[rgb]{0,0,0}\makebox(0,0)[lt]{\lineheight{1.25}\smash{\begin{tabular}[t]{l}$2$\end{tabular}}}}
    \put(0.72692767,0.09402702){\color[rgb]{0,0,0}\makebox(0,0)[lt]{\lineheight{1.25}\smash{\begin{tabular}[t]{l}$3$\end{tabular}}}}
    \put(0.80787928,0.0941838){\color[rgb]{0,0,0}\makebox(0,0)[lt]{\lineheight{1.25}\smash{\begin{tabular}[t]{l}$0$\end{tabular}}}}
    \put(0.88601092,0.09362118){\color[rgb]{0,0,0}\makebox(0,0)[lt]{\lineheight{1.25}\smash{\begin{tabular}[t]{l}$4$\end{tabular}}}}
    \put(0.64665174,0.01762651){\color[rgb]{0,0,0}\makebox(0,0)[lt]{\lineheight{1.25}\smash{\begin{tabular}[t]{l}$3$\end{tabular}}}}
    \put(0.7271495,0.01703982){\color[rgb]{0,0,0}\makebox(0,0)[lt]{\lineheight{1.25}\smash{\begin{tabular}[t]{l}$2$\end{tabular}}}}
    \put(0.80810112,0.01719688){\color[rgb]{0,0,0}\makebox(0,0)[lt]{\lineheight{1.25}\smash{\begin{tabular}[t]{l}$4$\end{tabular}}}}
    \put(0.88623276,0.01663426){\color[rgb]{0,0,0}\makebox(0,0)[lt]{\lineheight{1.25}\smash{\begin{tabular}[t]{l}$0$\end{tabular}}}}
    \put(0.47746014,0.41697478){\color[rgb]{0,0,0}\makebox(0,0)[lt]{\lineheight{1.25}\smash{\begin{tabular}[t]{l}$M_c:$\end{tabular}}}}
    \put(0,0){\includegraphics[width=\unitlength,page=5]{ted_calc_matrices.pdf}}
    \put(0.16363249,0.33143338){\color[rgb]{0,0,0}\makebox(0,0)[lt]{\lineheight{1.25}\smash{\begin{tabular}[t]{l}$1$\end{tabular}}}}
    \put(0,0){\includegraphics[width=\unitlength,page=6]{ted_calc_matrices.pdf}}
    \put(0.24369618,0.33119492){\color[rgb]{0,0,0}\makebox(0,0)[lt]{\lineheight{1.25}\smash{\begin{tabular}[t]{l}$2$\end{tabular}}}}
    \put(0,0){\includegraphics[width=\unitlength,page=7]{ted_calc_matrices.pdf}}
    \put(0.32245808,0.33186072){\color[rgb]{0,0,0}\makebox(0,0)[lt]{\lineheight{1.25}\smash{\begin{tabular}[t]{l}$3$\end{tabular}}}}
    \put(0,0){\includegraphics[width=\unitlength,page=8]{ted_calc_matrices.pdf}}
    \put(0.08129138,0.25287698){\color[rgb]{0,0,0}\makebox(0,0)[lt]{\lineheight{1.25}\smash{\begin{tabular}[t]{l}$1$\end{tabular}}}}
    \put(0,0){\includegraphics[width=\unitlength,page=9]{ted_calc_matrices.pdf}}
    \put(0.08084771,0.17322773){\color[rgb]{0,0,0}\makebox(0,0)[lt]{\lineheight{1.25}\smash{\begin{tabular}[t]{l}$2$\end{tabular}}}}
    \put(0,0){\includegraphics[width=\unitlength,page=10]{ted_calc_matrices.pdf}}
    \put(0.08129138,0.09357819){\color[rgb]{0,0,0}\makebox(0,0)[lt]{\lineheight{1.25}\smash{\begin{tabular}[t]{l}$3$\end{tabular}}}}
  \end{picture}
\endgroup
 }
			\caption{}
			\label{fig:ted_calc_matrices}
		\end{subfigure}
		\caption{(b) provides the $\sdted$ between pairs of $0$-unfolding trees ($M_r$) and $1$-unfolding trees ($M_c$) which are necessary to compute $\sdted(T,T')$.
		Following the order on node labels, resp. child trees, as in $M_r$, resp. $M_c$, the unfolding tree vectors of $T$ and $T'$ have the form $\bbV_r(T)=[1,0,0,0]$, $\bbV_c(T)=[1,1,0,1]$ and $\bbV_r(T')=[1,0,0,0]$, $\bbV_c(T')=[0,0,1,2]$. 
		One can check that $\W^{M_r}(\bbV_r(T),\bbV_r(T')) = 0$ and $\W^{M_c}(\bbV_c(T),\bbV_c(T')) = 4$, resulting in $\sdted(T,T') = 4$. }
		\label{fig:ted_calc}
	\end{figure}

	As an example, consider the $\sdted$ between graphs $T$ and $T'$ of Fig.~\ref{fig:ted_calc_trees}. 
	We assume that each insertion, deletion and relabeling operation has cost $1$.
	Following Def. \ref{def:1}, the root of $T$ is mapped onto the root of $T'$.
	As both vertices have the same label, the respective cost is zero (i.e. $\gamma(\ell(v_1),\ell(v'_1)) = 0$).
	Due to the structure preserving property of $\sdm$s, calculating the edit costs for the remaining vertices beneath the roots comes down to matching (resp. inserting and deleting) the highlighted subtrees. 
	It can easily be checked that matching $T[v_2]$ with $T[v'_2]$ (which has cost $2$) and thus deleting $T[v_3]$ (which has cost $2$) has minimal cost over all possible matchings. 
	The individual edit operations corresponding to this case are depicted in Fig.~\ref{fig:sdm_example_pos}.
	
	By the construction of unfolding trees, vertices closer to $v$ in $G$ begin to appear at smaller depths in $T^i(G,v)$.
	In fact, the number of occurrences in $T^i(G,v)$ of a node $u \in V(G)$ grows exponentially with $i$ once it has appeared for the first time. 
	This indirectly assigns higher weights to vertices closer to $v$ in the calculation of the structure and depth preserving tree edit distance.
	
	Notice that Algorithm \ref{alg:1} describes a naive implementation which in general requires an exponential number of recursion calls. 
	However, it is easy to see that the number of $i$-unfolding trees in $T$ and $T'$ is bounded by their sizes $n=V(T)$ and $n'=V(T')$. 
	Once $\sdted(T_i,T_j)$ between two $i$-unfolding trees $T_i,T_j$ has been calculated, it can be stored in a lookup table.
	Thus, for each level $i$, we need to invoke Algorithm~\ref{alg:1} a maximum of $nn'$ times.
	With a lookup table for distances between unfolding tree pairs we thus require at most $nn'h$ invocations of a minimum cost perfect bipartite matching algorithm, each of complexity $\tilde{O}((2d)^3)$, where $h$ is the depth and $d$ the maximum degree of $T,T'$. 

\section{The Relaxed Weisfeiler-Lehman Subtree Kernel}\label{sec:kernel}
	Using the definitions and results of Sect.~\ref{sec:editdistance}, we now introduce our novel \emph{relaxed Weisfeiler-Lehman subtree kernel} and show that it is in fact a \emph{generalization} of the original Weisfeiler-Lehman subtree kernel \citep{shervashidze2011weisfeiler}. 
	Its key idea is to \emph{relax} the rigid comparison of unfolding trees by equality (i.e., isomorphism) used in the Weisfeiler-Lehman kernel by considering the structure and depth preserving distances between unfolding trees. 
	Using $\sdted$, we identify groups of similar trees by means of \emph{hard} clustering.
	This ensures that similar unfolding trees will belong to the same clusters, while dissimilar to different ones.   
	Two unfolding trees are then regarded equivalent by the relaxed Weisfeiler-Lehman subtree kernel iff they belong to the same cluster.
	
	More precisely, for a set $\G$ of graphs, let $\Theta_i$ be a set of hard clustering functions (i.e., partitionings) of the set of depth-$i$ unfolding trees $\T^{(i)}$ appearing in the graphs in $\G$. 
	We regard each element of $\Theta_i$ as a function $\rho: \T^{(i)} \rightarrow [k]$, where $k$ is the number of clusters defined by $\rho$.
	Then, for any graphs $G,G' \in \G$ and depth parameter $h$, the {\em relaxed Weisfeiler-Lehman subtree kernel} is defined by
	\begin{gather*}
		k_\text{\sc R-WL}^h(G,G') = \\ \sum_{i=0,..,h} ~ \sum_{\rho \in \Theta_i} ~ \sum_{v \in V}\sum_{v' \in \ V'} \delta(\rho(T^i(G,v)), \rho(T^i(G',v')))   \enspace , 
	\end{gather*}	
	where $\delta$ is the Kronecker delta.
	Clearly, $k_\text{\sc R-WL}^h(G,G')$ is positive semi-definite and hence a kernel as the right hand side can be rewritten as the inner product of graph feature vectors  consisting of cluster membership counts (proof in Appenix A).
	Notice that $k_\text{\sc R-WL}^h$ is equivalent to the original Weisfeiler-Lehman subtree kernel $k_\text{\sc WL}^h$ for the case that $\Theta_i = \{\rho_i\}$ with $\rho_i$ defined as follows: For all $T,T' \in \T^{(i)}$, $\rho_i(T) = \rho_i(T')$ iff $T$ and $T'$ are isomorphic  (or equivalently $\sdted(T,T') = 0$).
	Thus, our definition generalizes the ordinary Weisfeiler-Lehman subtree kernel in two ways: First, while the ordinary Weisfeiler-Lehman subtree kernel regards two unfolding trees $T,T'$ to be equivalent iff $\sdted(T,T') =0$, our definition allows $\sdted(T,T') \geq 0$ as well. Second, our definition enables more than one partitioning (or hard clustering) function, in contrast to  $k_\text{\sc WL}^h$.

	We employ the concept of {\em Wasserstein $k$-means clustering}~\cite{DBLP:journals/eswa/IrpinoVC14} as a method to partition the set of unfolding trees.
	This choice is motivated by several arguments. 
	As mentioned above, the purpose of clustering is to group similar unfolding trees w.r.t. $\sdted$.
	We therefore require the clusters to be \emph{convex} such that unfolding trees of a cluster ideally have pairwise small distance.  
	Another requirement is to be able to control the number of clusters which also influences the complexity of the approximation variant of the relaxed Weisfeiler-Lehman kernel discussed in Sect.~\ref{ss:fasterkernel}. 
	We show that the $\sdted$ can in fact be calculated using the discrete Wasserstein distance. 
	Thus, we use the \emph{same} distance in the cost matrix as in the clustering process. 
	Finally, the Wasserstein distance has recently been the focus of comprehensive research leading to \emph{fast} approximation methods for distance and center computations \citep{DBLP:conf/icml/CuturiD14}.

	Below we address the most important ingredients of Wasserstein $k$-means for our purpose. 
	In particular, we first discuss how unfolding trees can be represented by real-valued vectors. 
	Subsequently, we state that the Wasserstein distance between such vectors corresponds to the $\sdted$ of the respective unfolding trees. 
	This description, furthermore, allows for the calculation of center points using {\em Wasserstein barycenters}. 
	For space limitations, we solely outline these concepts in this article. 
	A more detailed description as well as a complexity analysis can be found in the appendix.

	\paragraph{Unfolding Tree Vectors}
	\label{ss:unfoldingtrees}
		In order to effectively apply Wasserstein k-means, the unfolding trees need to be represented by real-valued vectors. 
		Recall that the structure and depth preserving tree edit distance is calculated as the sum of (A) the distance between the roots and (B) the minimum cost of a perfect bipartite matching between child trees below these roots (c.f. Alg. \ref{alg:1}). 
		We therefore represent an $i$-unfolding tree $T$ as a pair $\bbV(T)= (\bbV_r(T), \bbV_c(T))$, where the vector $\bbV_r(T)$ represents the root node's label $\ell(r(T))$ and $\bbV_c(T)$ represents the set of $(i-1)$-unfolding child trees $F(r(T))$.	
		$\bbV_r(T)$ is realized by a vector with entry $1$ at index corresponding to its root node label $\ell(r(T))$ and $0$ everywhere else, and $\bbV_c(T)$ is made up of counts of isomorphic child trees below the root. Analogously to Alg.~\ref{alg:1}, the vector $\bbV_c(T)$ furthermore contains an entry for empty child trees ($\bot$) to account for insertion and deletion. 
		An example of these vector representations is contained in the description of Fig. \ref{fig:ted_calc}.

	\paragraph{The Wasserstein Distance over Unfolding Tree Vectors}
	\label{ss:wasserstein}
		Using the vector representations of unfolding trees, we are able to reformulate the computation of the structure and depth preserving distance in terms of the Wasserstein distance. 
		Assume that the pairwise distances between child trees (as well as the empty tree) have already been calculated and are stored in a matrix $M_c$.
		Furthermore, let $M_r$ be the distance matrix between original node labels.  
		We can show that for two depth-$i$ unfolding trees $T$ and $T'$, the distance between their roots is equal to $\W^{M_r}(\bbV_r(T),\bbV_r(T'))$.
		Furthermore, the calculation of the minimum cost perfect bipartite matching between the sets of child trees below these roots (cf. Alg.~\ref{alg:1}) can be reduced to computing the Wasserstein distance between $\bbV_c(T)$ and $\bbV_c(T')$, i.e., $\W^{M_c}(\bbV_c(T),\bbV_c(T'))$.
		Putting all together we have:
		\begin{gather*} 
			\sdted(T,T') = \\
			\W^{M_r}(\bbV_r(T),\bbV_r(T')) + \W^{M_c}(\bbV_c(T),\bbV_c(T')) \enspace .
		\end{gather*}
		An example of this equivalence is given in Fig. \ref{fig:ted_calc}.

	\paragraph{Unfolding Tree Barycenters} 
	\label{ss:barycenters}
		The above reformulation allows us to calculate \emph{barycenters} of sets of unfolding trees for Wasserstein $k$-means. 
		A barycenter of a set $S$ of unfolding trees is a point which minimizes the sum of distances to unfolding tree vectors corresponding to $S$. 
		Similarly to unfolding tree vectors, this barycenter is a pair of real-valued vectors $(\mu_r,\mu_c)$, where $\mu_r$ is the center of the $\bbV_r$s and $\mu_c$ of the $\bbV_c$s. 
		More formally, the barycenter of $S$ is a pair $(\mu_r,\mu_c)$ calculated as follows: 
		\begin{equation} 
			\argmin\limits_{\mu_r, \mu_c} \sum\limits_{T \in S} \W^{M_r}(\bbV_r(T),\mu_r) + \W^{M_c}(\bbV_c(T),\mu_c) 
			\label{eq:barycenter}
		\end{equation}
		Note that while a barycenter, in general, does not correspond to an existing unfolding tree, the Wasserstein distance between an unfolding tree vector $\bbV(T) = (\bbV_r(T),\bbV_c(T))$ and a center vector $\mu = (\mu_r,\mu_c)$ can be computed nonetheless as follows:
		\begin{equation} 
			\W^{M_r}(\bbV_r(T),\mu_r) + \W^{M_c}(\bbV_c(T),\mu_c)
			\label{eq:wasserstein_distance}
		\end{equation}

	\paragraph{The Wasserstein $k$-Means Algorithm for Unfolding Trees}
	\label{ss:wassersteinkmeans}
		
		Using the above concepts, the application of the Wasserstein k-means clustering algorithm for unfolding trees is straightforward.
		(i) In the initialization step, a subset of $k$ unfolding trees is selected as initial centers.
		(ii) Each unfolding tree is then assigned to its nearest center point (using equation \ref{eq:wasserstein_distance}). 
		(iii) Finally, the centers of the newly defined clusters are recalculated (using equation \ref{eq:barycenter}). 
		Steps (ii) and (iii) are repeated until clusters do not change anymore, i.e., the algorithm converges, or a predefined number of iterations has been reached.

\subsection{A Faster Kernel Variant}
	\label{ss:fasterkernel}

	For many graph datasets the number of Weisfeiler-Lehman labels, or equivalently the number of (pairwise non-isomorphic) unfolding trees, grows rapidly with increasing iterations (although it is bounded by the total number of vertices in the database).
	Dealing with large amounts of unfolding trees is computationally expensive.
	We thus propose a variant of our kernel which approximates distances between unfolding trees using their cluster centers.
	
	Consider the calculation of pairwise distances between unfolding trees as in Sect.~\ref{ss:tedalgorithm}.
	That is, the distances of $0$-unfolding trees are defined by the metric $\gamma$ and the $\sdted$s for all pairs of $(i+1)$-unfolding trees are computed using distances of $i$-unfolding trees.
	To reduce the number of distinct $i$-unfolding trees $\T^{(i)}$ (or equivalently labels $\Sigma_i$), we perform a clustering $C_1,...,C_k$ of $\T^{(i)}$ with centers $\mu_1,...,\mu_k$ as in Sect.~\ref{ss:wassersteinkmeans}.
	We then effectively replace each $i$-unfolding tree with its cluster center $\mu_j$ and compute the distance between $i$-unfolding trees $T \in C_j, ~ T' \in C_{j'}$ by the distance between their cluster centers. 
	Subsequently, these distances are used in iteration $i+1$. 
	Hence, in contrast to the computation of $k_\text{\sc R-WL}^h(G,G')$, our kernel variant $k_\text{\sc R-WL*}^h(G,G')$ considers only $k$ labels instead of $|\T^{(i)}|$ labels in iteration $i$.

\section{Empirical Evaluation}\label{sec:experiments}

\newcommand{\WL}[0]{\text{WL}}
\newcommand{\WWL}[0]{\text{WWL}}
\newcommand{\PWL}[0]{\text{PWL}}
\newcommand{\RWL}[0]{\text{R-WL}}
\newcommand{\RWLA}[0]{\text{R-WL*}}
\newcommand{\GS}[0]{\text{GS}}
\newcommand{\SP}[0]{\text{SP}}
\newcommand{\ODD}[0]{\text{ODD-STh}}
\newcommand{\BL}[0]{\text{VE-Hist}}

	Below, we evaluate the predictive performance of our approach on a set of established as well as novel real-world datasets. 
	Our results show that our approach increasingly outperforms all considered competitor kernels with growing density of dataset graphs. 
	
	We note that in this short version, we limit the evaluation to the approximation kernel $\RWLA$ as discussed in Sect. \ref{ss:fasterkernel}.
	This choice was made due to the fact that while the $\RWL$ kernel is well applicable to sparse graphs such as molecules (see Appendix D), an explicit consideration of all unfolding trees may become computationally too expensive on more complex graphs.

	\subsection{Experimental Setup} 
	We compare our approach to a selection of graph kernels and provide a baseline method to put the performances into perspective.
	We consider the Weisfeiler Lehman subtree ($\WL$) kernel \cite{shervashidze2011weisfeiler} (with parameter $h \in [5]$), the graphlet sampling ($\GS$) kernel \cite{DBLP:journals/jmlr/Shervashidze2009graphlet} (with parameters $\epsilon=0.1$, $\delta=0.1$ and $k \in \{3,4,5\}$), the shortest-path ($\SP$) kernel \cite{borgwardt2005}, and the $\ODD$ kernel \cite{martino2012} (with parameter $h \in [4]$) using the implementation of \cite{siglidis2018grakel}.
	Furthermore, we include the recently published Wasserstein Weisfeiler-Lehman graph ($\WWL$) kernel \cite{NIPS2019_WassersteinWeisfeilerLehman} and the Persistent Weisfeiler-Lehman ($\PWL$) graph kernel \cite{Rieck19b}.
	In both cases, we select the depth parameter $h=5$.
	As a baseline method ($\BL$), we employ a simple histogram kernel over the set of edge and node labels. 
	In case of our relaxed Weisfeiler-Lehman kernel $\RWLA$, we choose the number of clusters $k=\sqrt{|\Sigma_i|}$ and perform a total of $3$ clusterings (i.e. $|\Theta_i|=3$), using depth parameter $h$ up to $4$ and cost $1$ for all relabeling, deletion and insertion operations. 
	This particular choice for $k$ is made in order to select the number of clusters relative to the amount of Weisfeiler-Lehman labels in each iteration as well as to significantly limit the computational complexity of the clustering. 
	The prediction performances are measured in terms of accuracy obtained by support vector machines (SVM) using a $10$-fold cross-validation.
	In each fold, a grid search is used to identify the optimal kernel parameters.
	We report the mean and standard deviation over $5$ such cross-validation repetitions. 
	Furthermore, runtimes can be found in Appendix D.

	\subsection{Datasets} \label{ss:data}

	We conduct experiments on the benchmark datasets IMDB-BINARY and REDDIT-BINARY containing subgraphs of online networks \cite{KKMMN2016}. 
	IMDB-BINARY consists of collaboration networks between actors/actresses each annotated against movie genres, whereas graphs in REDDIT-BINARY represent user interactions in discussion forums with graphs being annotated by the type of forum. 
	Furthermore, we provide a set of novel real-world benchmark datasets of varying size and density. 
	The datasets \emph{EGONETS}-$x$ contain ego network graphs extracted from four different social networks. 
	They contain increasingly larger and more dense ego networks with growing index $x$.
	Here, ego networks are subgraphs induced by a vertex's neighbors.
	Graphs within each dataset were randomly chosen from the set of all egonets but underlie size- and density-specific constraints to ensure that a simple count of nodes and edges is not sufficient for prediction tasks.
	The learning task is to assign an egonet to the network it was extracted from. 
	We provide detailed structural properties of all datasets in Appendix D.

	\subsection{Results \& Discussion}

\begin{table*}[t]
	\begin{center}
		\begin{tabular}{c | c | c | c | c | c | c}
			& IMDB-B. & REDDIT-B. & EGONETS-1 & EGONETS-2 & EGONETS-3 & EGONETS-4 \\ 
			\hline
			VE-Hist & $70.82\pm 0.53$ & $84.07\pm 0.28$ & $29.30\pm 2.84$ & $27.70\pm 2.64$ & $25.10\pm 2.56$ & $24.60\pm 2.10$ \\
			\hline 
			WL & $72.68\pm 1.20$ & $76.08\pm 0.66$ & $54.50\pm 0.94$ & $58.60\pm 1.29$ & $57.70\pm 2.08$ & $57.00\pm 1.77$ \\
			GS & $66.28\pm 1.00$ & $78.02\pm 0.58$ & $64.10\pm 1.98$ & $56.80\pm 1.20$ & $53.30\pm 2.86$ & $51.30\pm 1.96$ \\
			SP & $49.32\pm 0.62$ & $49.86\pm 1.96$ & $67.00\pm 1.46$ & $58.20\pm 1.04$ & $56.60\pm 2.90$ & $54.30\pm 3.49$ \\
			ODD-STh & $56.52\pm 1.86$ & $-$ & $32.80\pm 5.84$ & $37.90\pm 2.19$ & $34.80\pm 2.80$ & $34.40\pm 6.40$ \\
			WWL & $\mathbf{72.94\pm 0.59}$ & $-$ & $58.20\pm 1.60$ & $64.60\pm 1.08$ & $58.10\pm 2.38$ & $56.30\pm 1.92$ \\
			PWL & $72.58\pm 0.80$ & $78.82\pm 0.31$ & $56.95\pm 2.15$ & $56.85\pm 2.48$ & $50.35\pm 2.60$ & $48.70\pm 1.55$ \\
			\hline
			R-WL* & $72.16\pm 0.87$ & $\mathbf{87.02\pm 0.38}$ & $\mathbf{68.70\pm 1.15}$ & $\mathbf{69.30\pm 1.20}$ & $\mathbf{71.00\pm 3.54}$ & $\mathbf{77.50\pm 0.94}$ \\
		\end{tabular}
	\end{center}
	\caption{Classification accuracies and std. deviations for large network benchmark datasets in $\%$. Cells marked ``$-$'' indicate computations which did not finish within 24 hours. }
	\label{tbl:real_world}
\end{table*}

Table \ref{tbl:real_world} lists the classification accuracies for datasets containing graphs extracted from online networks. 
While there are no large discrepancies between our method and the best performing comparison kernels on datasets IMDB-BINARY, REDDIT-BINARY and EGONETS-1 (which all have an average node-to-edge ratio up to roughly $1:4$), the $\RWLA$ kernel considerably outperforms all others on the three remaining EGONETS datasets which contain significantly higher density graphs. 
The performance gap between the RWL* kernel and the best performing competitor becomes increasingly larger with a growing density in the dataset graphs, leading to an above $20\%$ accuracy difference. 
It is noteworthy that in case of the EGONETS datasets, already for depth $h=2$ nearly all unfolding trees (i.e. depth-$2$ unfolding trees) appear only once in the respective dataset. 
Thus, the original $\WL$ kernel is not able to profit from any structural information exceeding node degrees as graphs share almost no $i$-unfolding trees for $i \geq 2$. 
In contrast, our approach clearly improves upon this limitation by identifying similar unfolding trees. 

We, furthermore, evaluated our approach on traditional molecular datasets as well as synthetic benchmark datasets. 
We observed that while our approach does not prove to be advantageous on datasets containing mainly sparse and noise-free graphs such as molecular data, it soon outperforms all other considered kernels on datasets containing fuzzy and structurally diverse graphs. 
(For a detailed description see Appendix D.)

In summary, it is apparent that the original Weisfeiler-Lehman kernel is only suitable when there are only few different unfolding trees in the graphs of the dataset such that these graphs share sufficiently many labels in order to compute meaningful similarities.
Our method makes up for this drawback. 
Its ability to identify similar vertex neighborhoods leads to major increases in predictive performance on  datasets containing \emph{noisy} and \emph{structurally diverse} graphs.

\section{Concluding Remarks}\label{sec:conclusion}
	We introduced a generalization of the Weisfeiler-Lehman graph kernel which allows for finer similarity measures between Weisfeiler-Lehman labels.
	Our evaluation showed that this generalization improves upon a key weakness of the original Weisfeiler-Lehman graph kernel and outperforms state-of-the-art methods. 
	We stress that while we presented our relaxed Weisfeiler-Lehman graph kernel only for the {\em subtree} kernel, the generality of our approach allows its application to \emph{all} Weisfeiler-Lehman graph kernels.
	Furthermore, our results motivate several research questions.
	For one, alternative approaches to define meaningful similarities between labels might reduce expensive minimum cost perfect bipartite matching (or equivalently, Wasserstein) computations. 
	On another note, as the distance function $\gamma$ on initial graph labels can be defined by an arbitrary metric, the extension to attributed graphs is straightforward and promising. 

\bibliographystyle{abbrv}
\bibliography{references_brief}

\end{document}